\documentclass[10pt]{article}
\usepackage[margin=1in]{geometry}

\usepackage{graphicx}
\usepackage{listings}

\usepackage[version=4]{mhchem}

\usepackage{float,subcaption,caption} 

\usepackage{natbib}

\usepackage{booktabs}
\usepackage{array}

\begin{document}


\title{Domain Adaptable Prescriptive AI Agent for
Enterprise}

\author{Piero Orderique$^\dag$, Wei Sun*, 
Kristjan Greenewald*\\
$^\dag$MIT, *IBM Research}






%
%
%


\maketitle




\begin{abstract}
%
%
\noindent Despite advancements in causal inference and prescriptive AI, its adoption in enterprise settings remains hindered primarily due to its technical complexity. Many users lack the necessary knowledge and appropriate tools to effectively leverage these technologies.
This work at the MIT-IBM Watson AI Lab focuses on developing the proof-of-concept agent, PrecAIse,  a domain-adaptable conversational agent equipped with a suite of causal and prescriptive tools to help enterprise users make better business decisions.
The objective is to make advanced, novel causal inference and prescriptive tools widely accessible through natural language interactions.
The presented Natural Language User Interface (NLUI) enables users with limited expertise in machine learning and data science to harness prescriptive analytics in their decision-making processes without requiring intensive computing resources. We present an agent capable of function calling, maintaining faithful, interactive, and dynamic conversations, and supporting new domains.
\end{abstract}








\section{Introduction}
There is a growing demand in enterprise for prescriptive AI solutions, which focus on \textit{prescribing} actions to optimize specific objectives. To support the prescriptive AI pipeline, causal analysis and policy learning are essential due to their ability to provide actionable insights and guide decision-making processes. Specifically, causal inference allows for leveraging non-experimental datasets to quantify the impact of different actions while avoiding spurious correlations. Meanwhile, policy learning helps in deriving hyper-segmented strategies that are interdependent across segments, evolving based on changing market conditions, and ensuring continuous improvement in decision-making processes.


While powerful, prescriptive AI has faced several barriers to widespread enterprise adoption. 
Firstly, developing and implementing prescriptive AI solutions involves advanced machine learning knowledge including causal analysis and optimization. This complexity can be a barrier for many organizations that lack the necessary technical expertise. Secondly, many end users (such as sales managers, marketers, and operations leaders), when provided with AI-generated recommendations, lack the technical knowledge required to interpret the results and understand how to apply them to their specific business contexts.  

Traditionally, AI recommendations are delivered to  users either via (a) the technical team analyzing data and presenting solutions (e.g., Jupyter notebooks, Tableau) along with recommendations, which can be challenging for end users to apply to new use cases without the continuous support from the technical team, or (b) purpose-built GUI systems which typically offer only one-way interaction and are passive, not allowing for dynamic two-way interactions. As many tasks involve complex user input (e.g., conditions for conditional treatment effects or specifying policy constraints) and complex outputs (e.g., policies), developing such a GUI would require significant investment and likely pose a steep learning curve for users.

%

Motivated by these observations, PrecAIse (pronounced as ``precise", which stands for \textbf{Pre}scriptive \textbf{C}ausal \textbf{AI} \textbf{S}olution for \textbf{Enterprise}), an initiative spearheaded by IBM Research, aims to overcome these barriers by enabling users with limited AI background to leverage cutting-edge causal inference and prescriptive tools through an intuitive and user-friendly Natural Language User Interface (NLUI).
PrecAIse focuses on leveraging the power of Large Language Models (LLMs) to enable natural-language interaction with users. The vision is to empower an LLM-based agent to utilize these tools through a simple GUI comprising a text-based input and both text and figure-based output. LLMs naturally lend themselves to high levels of generalization, making it straightforward and easy for users to provide nuanced problem specifications and queries using natural language.
Furthermore, as many users may be unfamiliar with causal thinking and how to properly use these tools and interpret their outputs, the LLM framework is highly amenable to a more "hand-holding" user experience. The agent can guide the user through each step of the process, provide tailored explanations for all outputs, and answer user questions while offering suggestions for next steps.

An initial prototype of PrecAIse was introduced in \cite{sun2024presAIse}, where it was applied to an airline pricing use case, demonstrating its potential in enterprise settings. This prototype, built entirely with in-context learning (ICL), required considerable labor to tailor prompts for the enterprise use case and extensive prompt engineering. However, it suffers from unreliable performance 
when a user query slightly deviates from the few-shot examples provided in the prompt. Additionally, the hard-coded follow-up responses are repetitive, unnatural, and frequently contain hallucinations.
The current project's goal is to enhance the reliability and robustness of PrecAIse, focusing on designing a domain-adaptable system that can be applied to various enterprise contexts with minimal modifications.   
With this in mind,  we have re-designed several key components of PrecAIse, and incorporated the following enhancements to the agentic workflow for causal decision making. 
\begin{itemize}  
 \item To enable a domain-adaptable agent, we have created a fully automated  pipeline that allows users to set up the LLM agent with a new dataset with minimal effort. The pipeline works by taking the user
input (i.e.,  a short natural language description of the use case and a dataset), and auto-generating a system prompt, model configuration
files, and labeled queries needed for  LLMs to recognize the columns, and the user intent specific to the domain of interest. 
  \item We have added the ability to programmatically identify missing information and  utilized a mix of prompt injection and chain-of-thought techniques to create natural follow-up interactions, reduce hallucinations, and produce diverse yet domain-appropriate responses.  
  \item A generative User Interface that facilitate seamless, two-way communication with users by producing multimodal outputs, such as figures, plots, and other visual aids, ensuring comprehensive and understandable information delivery. 
  \item In addition to in-context learning, the current pipeline allows fine-tuning LLMs (including intent recognition and parameter extraction) for a more accurate function calling capability. We have also updated the memory module to improve the conversation flow.
\end{itemize}

In summary, we have developed an LLM-based agentic workflow that offers a robust, natural-language chat-based user interface for the complex prescriptive AI pipeline. This streamlined, automated process sets up the workflow for any desired dataset with minimal human input  (and only requires a few hours of GPU time if prompt tuning is used). Moreover, our approach is highly generalizable, paving the way for a new framework that provides business end users with reliable, chat-based access to nearly any data science workflow.

%


\section[PrecAIse Overview]{PrecAIse Overview}\label{ch:precAIse_Architecture}

\subsection{Prescriptive AI Pipeline}

We now briefly describe the backend tools that PrecAIse supports (more details can be found in \cite{sun2024presAIse}).  We assume we have access to an observational dataset that consists of historical actions $A$, covariates $X$ (may potentially be high dimensional), and outcomes $Y$. 

\textbf{Causal Analysis} It allows the user to understand and reason about the impact of the considered actions on the outcome of interest. Specifically, the tools from our prior work \cite{greenewald2021feature} are capable of: 
\begin{enumerate}
    \item Estimate the average causal effect of any action $A=a_1$ compared to any other action $A = a_2$. It helps to identify the action that is most effective for the entire population. 
    \item Estimate the conditional average effect of any action $A=a_1$ compared to any other action $A = a_2$, conditioned on certain conditions on $X$ being satisfied. It provides insights on how different actions work on subsets of your data (e.g. age cohorts, geographies). 
    \item Identify which covariates are {direct causal parents} of the outcome $Y$. This may inform the design of new actions seeking to improve these covariates, or may simply aid the building of intuition about the system.   
\end{enumerate}

\textbf{Policy learning} Once we gain insights from the data, it’s natural to ask how we can leverage these insights to take better actions.
 The tools from our prior work \cite{subramanian2022prescriptivetrees} focus on learning a set of optimal policies for a given objective. The policies are presented in a tree form for enhanced interpretability and the framework can handle a wide variety of user-specified constraints (e.g., budget, fairness). Specifically, 
\begin{enumerate}
    \item Learn optimal policies subject to global, inter-segment, and intra-segment constraints.
    \item  Allow users to perform ``what if" analysis by changing specific conditions. Finetune policies and compare different counterfactual outcomes.
\end{enumerate}

\begin{center}
    \begin{table}
    \caption{Supported Prescriptive AI Tools}
    \label{fig:tools}
    \begin{tabular}{|p{4cm}|p{3cm}|p{2cm}|p{6cm}|}
    \hline
    \textbf{Tool Name} & \textbf{Parameters} & \textbf{Returns} & \textbf{Description} \\ 
     \hline
     show\_current\_policy & None & float, html bar graph & Shows what the current policy is and any relevant KPIs. \\
     \hline
     select\_features & None & list, html figure & Covariate selection tool that selects most the important features that affect the outcome. \\
     \hline
     show\_causal\_effect & show\_error & html plot & Plots how the action affects the outcome in the average case. \\
     \hline
     counterfactual & conditions & html plot & Plots how the action affects the outcome under the provided conditions. \\ 
     \hline
     run\_optimize & num\_rules, average\_budget & float, html bar chart, html tree graph & Produces the optimized KPI and policy through a prescriptive tree constrained by an average budget per row. \\
     \hline
    \end{tabular}
    \end{table}
\end{center}

\subsection{PrecAIse Architecture}

\begin{figure}[H]
    \centering
    \includegraphics[width=1\linewidth]{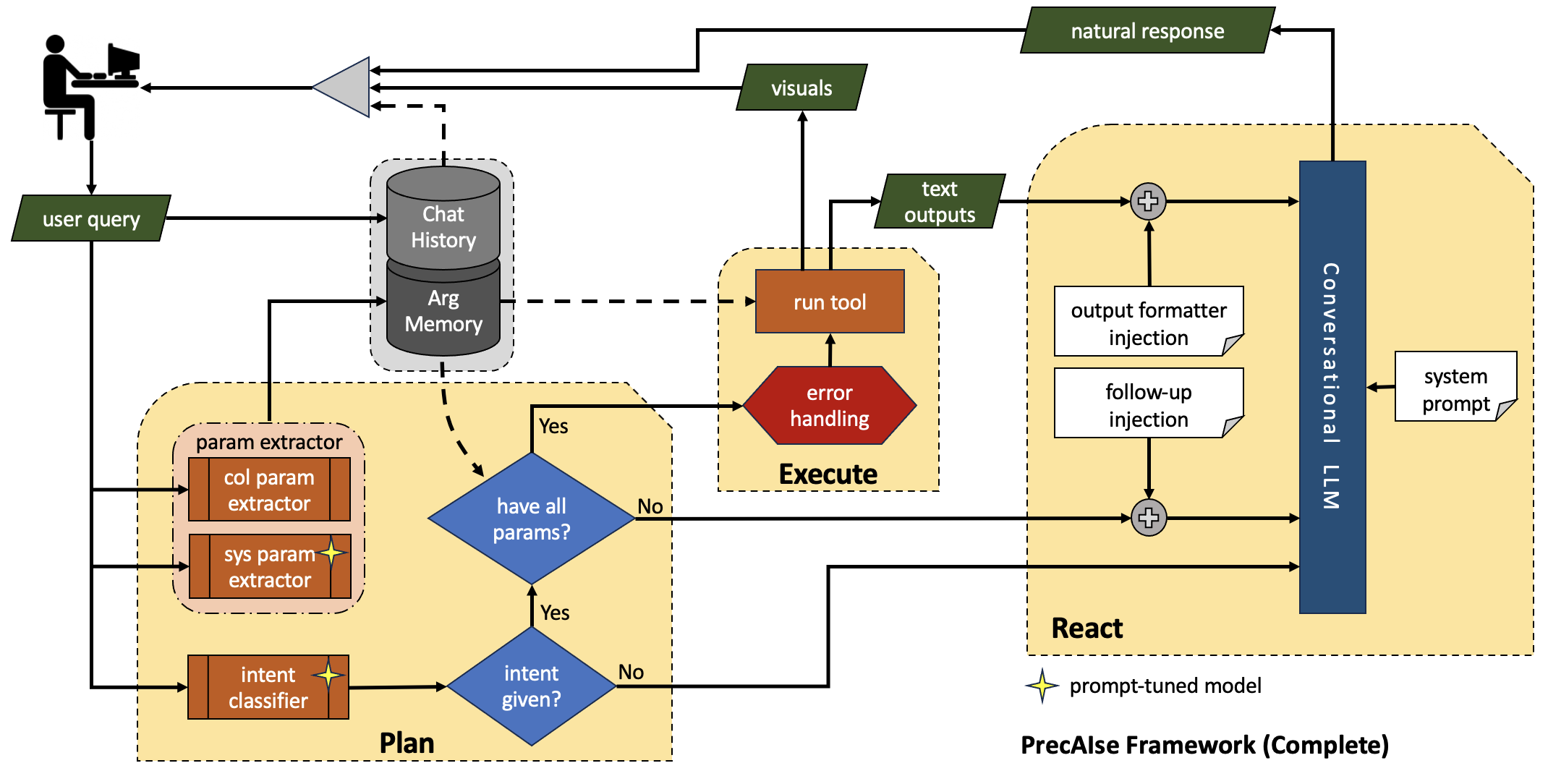}
    \caption{Complete Agent Framework}
    \label{fig:agent-framework}
\end{figure}

The  PrecAIse framework is presented in Figure \ref{fig:agent-framework} building on the architecture proposed in Sun et al. \cite{sun2024presAIse}. This work focuses on designing and implementing each module - specifically the function calling modules and conversational modules. This framework incorporates the updated tool wrappers from Table \ref{fig:tools}, prompt tuned models specialized in parameter extraction and intent classification introduced in Section \ref{ssec:prompt_tuning_method}, a new chat model, memory module, and thought injection techniques from Section \ref{sec:convo_methods}, and an updated system prompt shown in Figure \ref{fig:system-prompt}.

\begin{figure}
    \centering
    \includegraphics[width=0.75\linewidth]{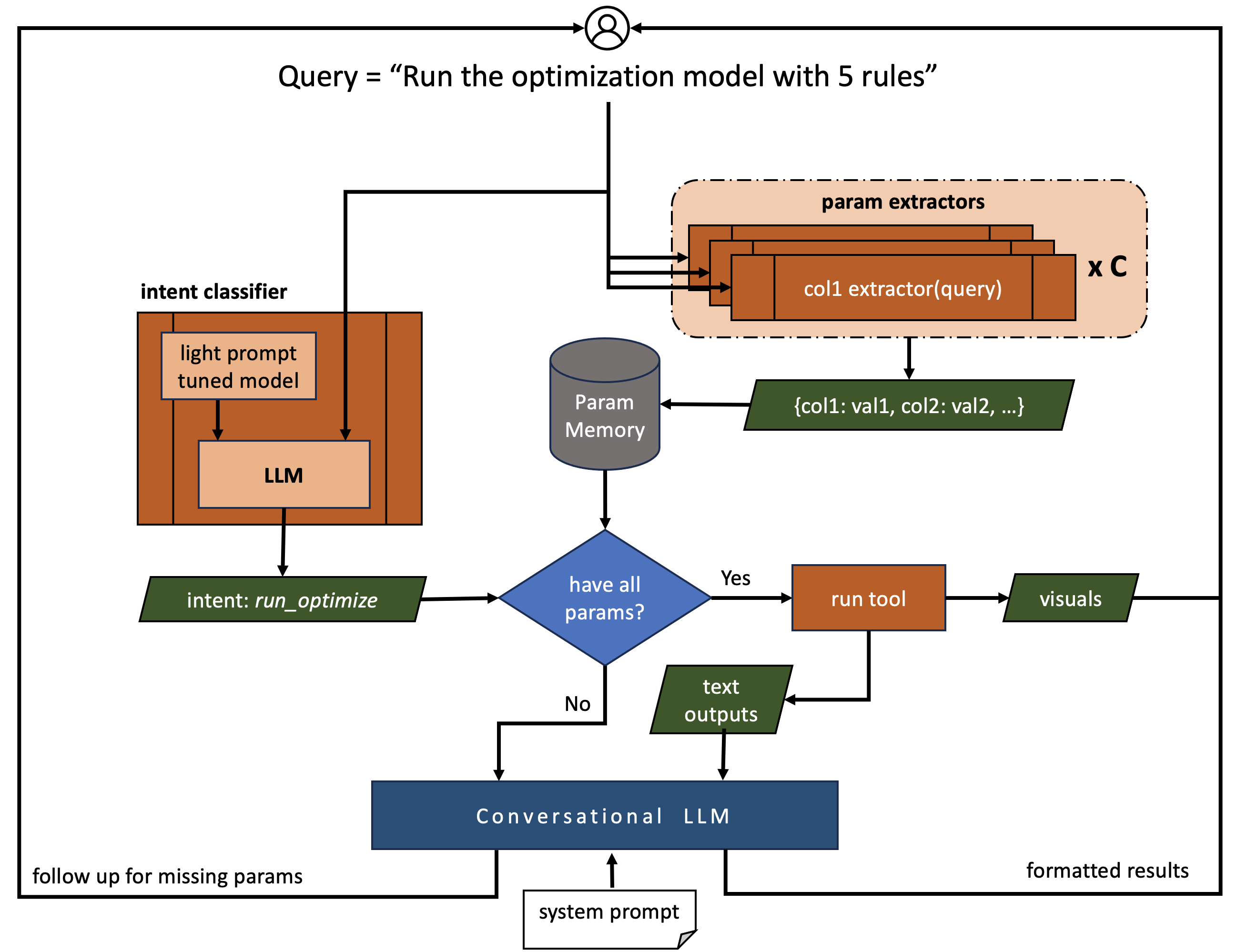}
    \caption{Updated Query Flow}
    \label{fig:agent-logic}
\end{figure}

\textbf{Logical Flow Walk-through.} Figure \ref{fig:agent-logic} shows the logical flow of a user's query in the system. The first part of the flow sends the incoming query to an intent classifier and several parameter extractors. These classifiers/extractors are prompt tuned for increased accuracy, which means that the original LLM weights are frozen and the input query is combined with a light, trained embedding. The intent classifier is responsible for mapping the query to a predefined tool name such as those in Table \ref{fig:tools}.

Simultaneously, the same query goes through all $C$ column extractors, each one responsible for extracting out the value corresponding to its specified column. These extractors are also prompt tuned like the intent classifier. This column to value mapping is then stored in parameter memory for future use.

Once the system knows the intended tool, it reads from the parameter memory to check if it has all the information needed to run the tool. If it does not, then the conversational LLM, equipped with a system prompt, is prompted to follow up for the missing parameters. If the system has all the parameters needed to run the tool, then the tool will be executed. Any textual outputs are passed through the conversational LLM to be formatted in natural language via thought injection, and visual outputs like html plots are passed back to the user directly.

\paragraph{Need for Generalization.} 
While this framework works well for a predefined dataset with specialized modules capable of supporting domain-specific queries from the user, a crucial part of this work is to be able to generalize to new datasets. Prior to this work, PrecAIse only worked as an airline pricing tool with no generalization and low-quality tool routing functionality. Much of the agent's functionality was rigid and hard-coded to fit the needs of the domain. For example, the tools used by the agent required parameters specific to the airline case like "origin" or "destination" inputs. As already introduced in Table \ref{fig:tools}, these tools have since been updated to handle generic parameters and their implementations have been decoupled from the airline use case. This new framework, however, introduces innovative modules and a generalization pipeline in Chapter \ref{ch:domain-adapt} that is able to transform PrecAIse into a general-purpose tool for bringing causal and prescriptive tools to the masses.

\section[Creating a Domain Adaptable Agent]{Creating a Domain Adaptable Agent}\label{ch:domain-adapt}

There are three modules that need to be updated to apply the agent to a new domain - the intent classifier, column extractors, and the system prompt. The updated intent classifier must be able to handle domain-specific queries and map them to one of the underlying tools. Similarly, there needs to be new column extractors to handle column-specific queries regarding the new dataset that is uploaded. Finally, the system prompt needs to take in some general information about the dataset to help guide the user properly.
\begin{figure}
    \centering
    \includegraphics[width=1\linewidth]{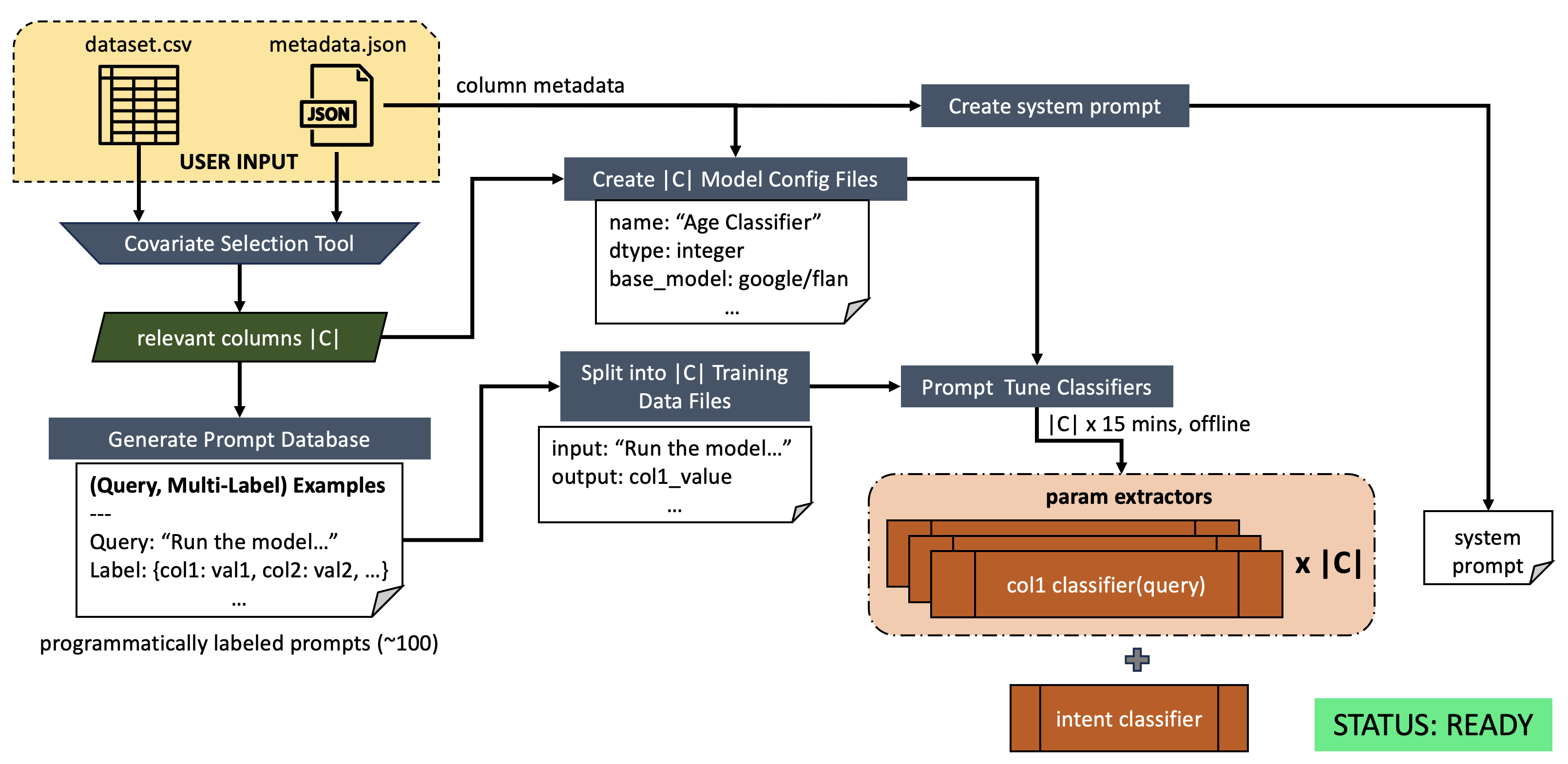}
    \caption{Automated Generalization Pipeline}
    \label{fig:gen-pipeline}
\end{figure}

We introduce a fully automated generalization pipeline in Figure \ref{fig:gen-pipeline} that is able to auto-create these two modules from simply the new dataset and some metadata about it. The pipeline works by taking the user input, selecting a subset of the dataset columns, and auto-generating a system prompt, model configuration files, and labeled queries needed for prompt tuning the column extractors and the intent classifier.

\textbf{User input.} The dataset itself is expected to be in comma-separated values (CSV) format. The metadata should contain fields about the dataset's title, path, action variable, outcome variable, and information about each column like its datatype and short description. This type of metadata is common for datasets to have, making it simple for users to get started.

\textbf{Covariate selection tool.} Using the dataset and its metadata, the covariate selection tool developed in Greenewald et al. \cite{greenewald2021feature} selects the most relevant columns in the dataset that, along with the action column, affect the outcome. This is a crucial step as it helps reduce the number of column extractors that need to be trained. If the client needs more flexibility, columns can be specified in the metadata file to be added/removed from these supported columns.

\textbf{Generating the prompt database.} Once $C$ columns are selected from the covariate selection tool, a database of $\approx100$ example (query, multi-label) samples are generated. ``Query'' is an example query we expect to receive from the user, and ``multilabel'' is a length $C$ mapping of each column name to the expected value to be extracted from the query plus a mapping for the expected intent classification and any other system parameters that should be extracted. This database consists of some hard-coded, domain-agnostic manual samples and domain-specific templates that are formatted using information from the database. 

An example of these queries can be seen in Table \ref{fig:example_queries}, where the domain-specific query templates will get the ``[ACTION]" and ``[OUTCOME]" variables from the metadata file, and will fill in ``[COLUMN]" placeholders with a chosen column $c \in C$, and the ``[VALUE]" placeholders with a value $v$ within that column $c$. The multi-label for this query is then updated with this new pair mapping $\{c: v\}$.

\begin{figure}
    \centering
    \begin{tabular}{|l|l|}
    \multicolumn{1}{c}{\bfseries Domain-Agnostic Queries} & \multicolumn{1}{c}{\bfseries Domain-Specific Queries}\\ 
    \hline
        What is the best action? & How does [ACTION] affect [OUTCOME]? \\ \hline
        Show the causal effect & What if [COLUMN] is [VALUE]? \\ \hline
        What are the most important features? & Is [COLUMN] an important variable? \\ \hline
    \end{tabular}
    \caption{Domain Agnostic vs Domain Specific Samples}
    \label{fig:example_queries}
\end{figure}

\textbf{Splitting into training data.} To train the intent classifier and column extractors, each model needs a training file consisting of (input, output) samples. For each column name $c$ in $C$, its training file is created by using every query in the prompt database as the inputs, and indexing into the multi-labels for that specific column's expected values to use as outputs.

\textbf{Creating model configuration files.} At the same time as the prompt database is being generated, the selected $C$ columns and their corresponding metadata is then used to create default model configuration files. These files include information necessary for creating the prompt tuned classifiers such as the datatype, the base model, and the initial text initialization. These defaults can of course be edited by the user if need be, but it is not required nor necessary in most cases. 

As discussed in Section \ref{ssec:prompt_tuning_method}, early observations showed that starting from a text-initialization, compared to a random initialization, produced much faster convergence time during training. The text initialization template used in the generalization pipeline for column extractors is shown in Figure \ref{fig:instruction-template}, where each missing field is either auto-generated or retrieved from the column metadata. Notice that it looks very similar to the instruction one would give for in-context learning.

\begin{figure}[htbp]
    \centering
        \begin{minipage}{1.0\textwidth}
            \begin{verbatim}
From each command given, extract out the value of "{param}" if specified.
Only output values corresponding to the datatype {dtype}.
If none is given or you are not sure, output {dtype_to_null_defaults[dtype]}.

{param} description:
{description}

<examples>
            \end{verbatim}
        \end{minipage}
    \caption{Instruction template used for text-initialization in prompt tuning column extractors}
    \label{fig:instruction-template}
\end{figure}

\textbf{Prompt tuning.} Once the system has generated both the configuration files and the training samples for each column (and for the intent classifier), the system will upload these files and query each model to be prompt tuned. Using internal IBM infrastructure tools, we found that tuning usually takes $15$ minutes using $100$ generated training samples. However, this is done offline - not requiring the user to be in the loop. After tuning, these models are ready to be deployed.

\textbf{System prompt.} The final missing piece is to create the system prompt. This prompt contains useful information about the dataset and outlines the goals and initiatives of the agent. The template is presented in Figure \ref{fig:system-prompt}. Once this is auto-generated and all the classifiers/extractors have been tuned, the agent is ready to be deployed.

\section{Methodology}

This section outlines the methods used to improve the current function calling and conversational abilities of the agent. Chapter \ref{ch:eval} discusses the notable quantitative and qualitative results from these changes.

\subsection[Improving Function Calling]{Improving Function Calling}
The pre-existing prototype of PrecAIse from Sun et al. \cite{sun2024presAIse} used few-shot learning to create an intent classifier and several parameter extractors. Each classifier/extractors would be given an instruction along with a set of manually written (query, expected output) samples. For example, PrecAIse used a few-shot prompt to extract out the price range in a given query and another few-shot prompt, with its own set of examples, to extract the origin and destination pair. If no value was identified in the query, the classifier/extractors would output ``unknown''. An example few-shot prompt for the price-range classifier is included as Figure \ref{fig:few_shot_prompt}.

\begin{figure}[htbp]
    \centering
        \begin{minipage}{1.0\textwidth}
            \begin{verbatim}
Extract the lower and upper price bound as numbers from Command.
If a price bound appears as a dollar value, omit the dollar sign.
The default lower price bound is 0, the default upper price bound is 1000.
All answers must be in the form "<min_price>-<max_price>".
Answer MUST INCLUDE the '-' character.
If a number cannot be extracted, output "Unknown"

<examples>
command: Change the maximum price to be 800
Unknown-800

command: Increase the lowest price to $200
200-Unknown

...
            \end{verbatim}
        \end{minipage}
    \caption{Original Few-Shot Prompt for Price Range Classifier}
    \label{fig:few_shot_prompt}
\end{figure}

\subsubsection[Early Observations and ICL Experiments]{Early Observations and ICL Experiments} 
This approach of having each classifier/extractor have its own set of individualized samples proved problematic. Queries such as ``Show me the policy for the BOS-ATL market'' would cause the price range classifier to output ``lower\_bound: BOS'' and ``upper\_bound: ATL''. This happens because in the few-shot examples provided to the price-range classifier, an origin-destination pair was never included; therefore, the model learns to treat anything in the format ``data-data'' as a price range. Furthermore, classifiers in charge of outputting numbers would often output whole words other than ``Unknown''.

A series of prompting techniques were explored to improve classification results. These included restricting outputs to one data type and including samples used for one classifier in the example set for other classifiers.

It was hypothesized that allowing numerical classifiers to output the word ``Unknown'' could confuse the model into thinking that non-numerical text in general was allowed. To help remove this confusion, we restricted numerical classifiers to output only numbers, and assigned them a numerical ``null'' value. For example, the price-range classifier's instruction was changed to output null values of ``1'' and ``9999'' as the default min and max values respectively. Furthermore, the examples themselves were changed by appending ``Price Range (\$):'' to the end of the prompt to signal to the model that a numerical pair needs to be returned.

While these updates qualitatively improved the classifier, the best results were observed when mixing few-shot examples from both the price range and origin-destination classifiers together. This way, the price range classifiers knew that queries containing market pairs, which is common in an airline case, have nothing to do with price ranges. This improvement led to the idea of using a centralized prompt database and keeping the few-shot examples the same across all the classifiers.

\subsubsection[Prompt Tuning]{Prompt Tuning}\label{ssec:prompt_tuning_method}
To achieve a professional level of engagement and usability, we decided to  fine-tune models to enhance the accuracy of the agent. We opted for parameter efficient fine tuning methods that kept model weights frozen in order to keep costs and inference times low. We used prompt tuning \cite{ibm-martineau} due to IBM's strong infrastructure support for this method.

To prompt tune a model, the creator needs to supply a series of standard hyperparameters (gradient accumulation steps, learning rate, number of virtual tokens, etc.) and a training file of (input, output) samples. The training examples are the same as the examples used in few-shot learning. For the hyperparameters, these defaults were chosen:
\begin{itemize}
    \item Gradient accumulation steps: 16
    \item Initialization method: text
    \item Learning rate: 0.3
    \item Number of virtual tokens: 500
\end{itemize}

The hyperparameter that had a noticeable impact on performance and training speed was the initialization method, which sets whether or not to start from a random or an initial text initialization. Starting from a random initialization means the soft prompt starts from a random embedding before tuning starts. Using text initialization, however, the process starts from the embedding of the passed in text. Since the few-shot classifiers being used already included an instruction string, this instruction was used as the text initialization. Figure \ref{fig:intent-prompt} shows the instruction used for the few-shot intent classifier that was reused as the initial text for the prompt tuning process.

\begin{figure}[htbp]
    \centering
        \begin{minipage}{1.0\textwidth}
            \begin{verbatim}
Classify command as one of following API calls. 
If none can be matched, just output unknown.
'select_features' cross validation plot of the causally relevant features
'show_causal_effect' shows the causal effect conditioned on given features.
'run_opt' produces the optimized pricing policy for given conditions.
'show_current_policy' shows the historical policy for given conditions.
'counterfactual' predicts the counterfactual outcome when columns are fixed.

If the intent is unclear, output `unknown'
            \end{verbatim}
        \end{minipage}
    \caption{Intent Classification Prompt}
    \label{fig:intent-prompt}
\end{figure}

\begin{figure}[htbp]
    \centering
    \includegraphics[width=0.75\linewidth]{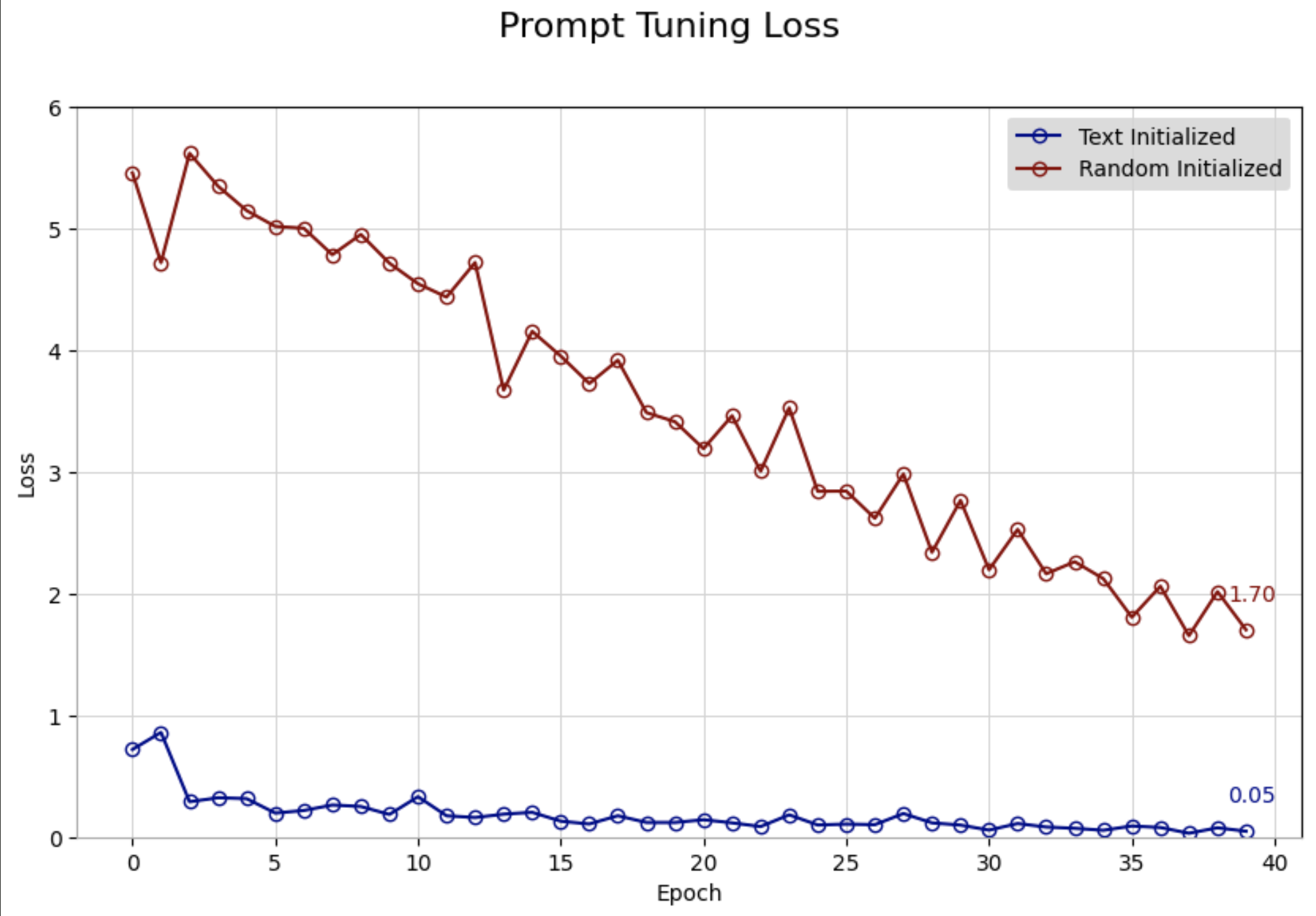}
    \caption{Text Initialization vs Random Initialization for Prompt Tuning}
    \label{fig:pt_loss}
\end{figure}

Figure \ref{fig:pt_loss} shows the loss graph for an intent classifier trained with all the same hyperparameters and training data for $40$ epochs each. Notice how the tune that started with the text initialization not only ended with a much smaller loss, but also started at a much lower loss than random initialization. The text used to initialize the soft prompt was the instruction used for the intent classifier's few-shot prompt.

Using this prompt tuning method, we were able to save resources, time, and improve the performance of these classifiers without significantly increasing the amount of work needed from writing few-shot prompts. Starting from this initial text ensures that the setup process can be done more efficiently once the agent is adapted to support new domains as discussed in Chapter \ref{ch:domain-adapt}. The performance of the prompt tuned intent classifier is discussed in Section \ref{eval:function-calling}, however, the parameter extractors were also prompt tuned in order to achieve noticeable performance gains.

\subsection[Improving Conversational Experiences]{Improving Conversational Aspects}\label{sec:convo_methods}
The pre-existing PrecAIse prototype handled conversations by also using a few-shot prompt. This resulted in rigid responses and did not have the capacity to follow up for missing information. We were able to significantly improve the agent's conversational abilities, including the capacity to engage in natural conversations, follow up, and present non-hallucinated information. Qualitative results across these dimensions are shown in Section \ref{eval:convo}.

\textbf{Models.} Rather than using a few-shot prompt to create the conversational LLM, we opted for a  Sparse Mixture of Experts model developed by the Mistral AI team \cite{mixtral}. More specifically, we used the "Mixtral 8x7B Instruct" model, that was fine-tuned on instruction based tasks in order to have more fine-grain control over the model's output. To have it behave as a chat model, the model was simply instructed to do so using a system prompt commonly seen in chat based models. The full system prompt used can be seen in Figure \ref{fig:system-prompt}.

\begin{figure}[htbp]
    \centering
        \begin{minipage}{1.0\textwidth}
            \begin{verbatim}
system:
You are a friendly and cheery AI agent named PrecAIse, pronounced `Precise'.
Your job is to assist analysts to determine the optimal policy.
You were built with a goal to help business users make better decisions by
leveraging the power of AI.

You are working with prescriptive policy models using a {TITLE} dataset.
Action variable is {ACTION}. 
Outcome is {OUTCOME}.

Based on every user's query, you identify their intent from the following:
- select_features
- show_causal_effect
- run_opt
- show_base_policy
- counterfactual

The key functionalities that you currently support include 
- selecting the important features for treatment effect estimation
- quantifying the treatment effect
- quantifying the treatment effect conditioned on covariate values
- generating a set of optimized policies
- evaluating the KPIs
- and predicting counterfactual scenarios.

When a user query can be mapped to one of the existing functionalities 
with necessary parameters, reply with enthusiasm that you are happy to 
assist the user and you are working on the query.

You are harmless and refrain from generating content involving any form 
of bias, violence, discrimination or inappropriate content. 
Do not say anything outside the field of the dataset and 
prescriptive analysis and do not start a conversation off topic to causal 
inference. 

If prompted off topic or given a silly request, kindly redirect user 
back to the task at hand.
If the user is asking for a tool to be used, tell them you're happy to 
help with that. Always keep responses as short (under 50 words) and 
concise as possible and only expand when prompted. 
Do not make up information.
            \end{verbatim}
        \end{minipage}
    \caption{System prompt template}
    \label{fig:system-prompt}
\end{figure}

\textbf{Chat Memory.} The agent's chat memory was updated to incorporate both semantic and short-term memory. The agent, as before, still stores previously extracted parameters so that it does not have to follow-up for this information later on. Now, however, the agent is also capable of remembering recent interactions to allow for a more natural feel without incurring significant speed costs. 

A design choice here is to decide how many $k$ recent interactions to store in the context. A higher $k$ leads leads to higher inference times as the context passed in to the LLM gets larger, however it allows the model to remember more information. Another drawback noticed in practice is that given that the user is usually running multiple tools one after the other, an agent with high $k$ tends to treat its own memory of previous responses as few-shot examples, leading to the same rigid answer structures seen with using ICL methods. For this reason,  $k=2$ was used since in practice it seemed to be a good balance of recent interactions without compromising its ability to generate diverse, natural responses.

\textbf{Hallucination.} Since this agent is intended for professional use cases, it is imperative that responses are tailored on real data and that the model does not hallucinate effects, policies, or results. The previous version of PrecAIse suffered from making up information that seemed relevant to the context application, but in reality was based on false or hallucinated analyses. To help mediate this, we introduce a deviation of prompt injection \cite{prompt_injecttion}. In the basic case where the agent needs to provide a response to a users query, the chat model is prompted as shown in Figure \ref{fig:chat-prompt}.

\begin{figure}[htbp]
    \centering
        \begin{minipage}{1.0\textwidth}
            \begin{verbatim}
*system prompt inserted here*

<Chat History>
USER: Hello how are you?
AGENT:
            \end{verbatim}
        \end{minipage}
    \caption{Basic Chat Prompt}
    \label{fig:chat-prompt}
\end{figure}

In scenarios where a tool has been executed, we inject a mini system prompt right before prompting for the agent's response. This works similarly to chain-of-thought prompting, but instead this part of the "chain" is a programmatically created command instructing the agent how to respond. This method works especially well since we use the instruction fine-tuned version of the Mixtral 8x7B model. This injected command reduces the chances of hallucination while still allowing the model some freedom to generate a diverse response grounded in results (rather than a hard-coded template each time). This "thought injection" method is used to instruct the model to follow up for missing parameters or respond back to the user with tool results as shown in Figures \ref{fig:followup-injection} and \ref{fig:output-injection}, respectively, with the "SYSTEM" tag.

To further improve interpretability, this thought injection technique was also used to inform the user what tools are being run in the background (along with the parameters that were extracted) as seen in Figure \ref{fig:tool-injection}. This ensures that the agent has some credibility instead of simply presenting results from tool execution.

\begin{figure}[htbp]
    \centering
        \begin{minipage}{1.0\textwidth}
            \begin{verbatim}
*system prompt inserted here*

<Chat History>
USER: Can you optimize my strategy?
SYSTEM: Respond to the users query but ask to provide the following
    missing parameters: [num_rules, avg_budget]
AGENT: 
            \end{verbatim}
        \end{minipage}
    \caption{Thought Injection: Follow up for missing parameters}
    \label{fig:followup-injection}
\end{figure}

\begin{figure}[htbp]
    \centering
        \begin{minipage}{1.0\textwidth}
            \begin{verbatim}
*system prompt inserted here*

<Chat History>
USER: What is the current conversion rate?
SYSTEM:  Simply respond to the user that the result is <INSERT RESULT>.
    Say nothing else and do not make up anything.
AGENT: 
            \end{verbatim}
        \end{minipage}
    \caption{Thought Injection: Present tool outputs}
    \label{fig:output-injection}
\end{figure}

\begin{figure}[htbp]
    \centering
        \begin{minipage}{1.0\textwidth}
            \begin{verbatim}
*system prompt inserted here*

<Chat History>
USER: How does the campaign column affect conversion?
SYSTEM: Inform the user you are running a tool that does
    <insert tool description>
AGENT: 
            \end{verbatim}
        \end{minipage}
    \caption{Thought Injection: Provide backend insights}
    \label{fig:tool-injection}
\end{figure}

Refer to Section \ref{eval:convo} for qualitative comparisons of answers before and after the improvements made in each area.

\section{Evaluation}\label{ch:eval}

\subsection[Evaluating Function Calling]{Evaluating Function Calling}\label{eval:function-calling}
To evaluate the agent's function calling ability, we focus mainly on its capacity of recognizing the user's intent and mapping it to one of the predefined functions listed in Table \ref{fig:tools}. The following is the developed testing framework to quantitatively asses the agent's function calling abilities. We show that prompt tuning the intent classifier using this framework results in a better model than using ICL methods without introducing any additional manual work than is already needed to support few-shot learning.

\textbf{Manual Training Examples.}
We map out initial experiments using a manual database of $\approx50$ (query, tool) examples as the "training data" for ICL and prompt tuning methods. In the the ICL case, these are used as the few-shot examples prompted to the model, while in the prompt tuning case, these are used as training examples to update the weights of the learned soft prompt.

\textbf{Programmatically Generated Training Examples.}
Due to the need to generalize this pipeline, as discussed in Chapter \ref{ch:domain-adapt}, we created templates from the manual training examples and used information directly from the columns of the database to fill in the blanks. Moreover, doing so allowed us to double the amount of training examples used for both ICL and prompt tuned models.

\textbf{Creating Synthetic Test Data.} Finally, the evaluation pipeline needs ample amounts of test data to get decent insights to these classifier's performances. While prompting based approaches have been able to effectively generate labeled data for intent classification \cite{sahu2022dataaug}, these methods focus on generating diverse utterances with a similar intent as the given query. This means that 1) this process would have to be run multiple times since there are multiple queries, 2) out-of-domain queries might slip into the testing set, and 3) no other needed labels, i.e. function parameters, are involved, making this process seem less feasible in this use case. Even though using these prompting-based approaches with a low temperature has shown to improve generation \cite{holtzman2019temperature}, controlling for out of domain utterances would still introduce manual oversight.

For this reason, we utilized a work called Parrot, a paraphrase-based utterance augmentation framework \cite{parrot}. Parrot works by preserving the meaning of a sentence while ensuring fluency and diversity, thus enabling a nuanced evaluation of the agent's function calling proficiency. Using Parrot on the the $\approx50$ manual examples we had created earlier, we generated $238$ faithful, synthetic examples for testing.

\textbf{Performance.}
Table \ref{fig:intent-eval} shows performance metrics for ICL and prompt tuned models trained on either the manual or generated training examples. All were tested on the $238$ synthetically generated queries.

\begin{center}
    \begin{table}[ht]
    \caption{ICL vs prompt tuned models}
    \label{fig:intent-eval}
    \begin{tabular}{|p{4cm}|p{2.2cm}|p{2cm}|p{2cm}|p{1.5cm}|p{2cm}|}
    \hline
    \textbf{Model} & \textbf{Accuracy} & \textbf{F1 Score} & \textbf{Precision} & \textbf{Recall} & \textbf{Inference Time}\\ 
     \hline
     ICL w/ manual training set & 0.64 (152/238) & 0.55 & 0.61 & 0.64 & 0.58s\\
     \hline
     Prompt Tuned w/ manual training set & 0.88 (209/238) & 0.82 & 0.80 & 0.87 & 0.45s\\
     \hline
     ICL w/ generated training set & 0.63 (151/238) & 0.61 & 0.62 & 0.65 & 0.66s \\
     \hline
     Prompt Tuned w/ generated training set & \textbf{0.95 (226/238)} & \textbf{0.94} & \textbf{0.94} & \textbf{0.94} & \textbf{0.44s}\\
     \hline
    \end{tabular}
    \end{table}
\end{center}

Using the generated training set allowed for more training examples to be used and improved the f1 score, precision, and recall in both the ICL and prompt tuning models. Since these are being derived programmatically from set templates, it helps validate that assumption that these templates will work well in producing better classifiers when the agent is applied to a new domain with a new dataset, as done in Chapter \ref{ch:domain-adapt}.

Notice that the inference time increases in the ICL case from 0.58s to 0.66s when switching from the manual examples to the generated ones. This is expected behavior since the generated examples includes double ($\approx100$) the amount of examples than the manually written ones ($\approx50$) and so the model has a longer context on inference time. However, the inference time of the prompt tune models stays relatively the same - going from 0.45s to 0.44s - when switching from the manual to the generated test cases. This is because the size of the learned prompt stays fixed no matter how many training examples are seen.

\begin{figure}[H]
    \centering
    \includegraphics[width=0.6\linewidth]{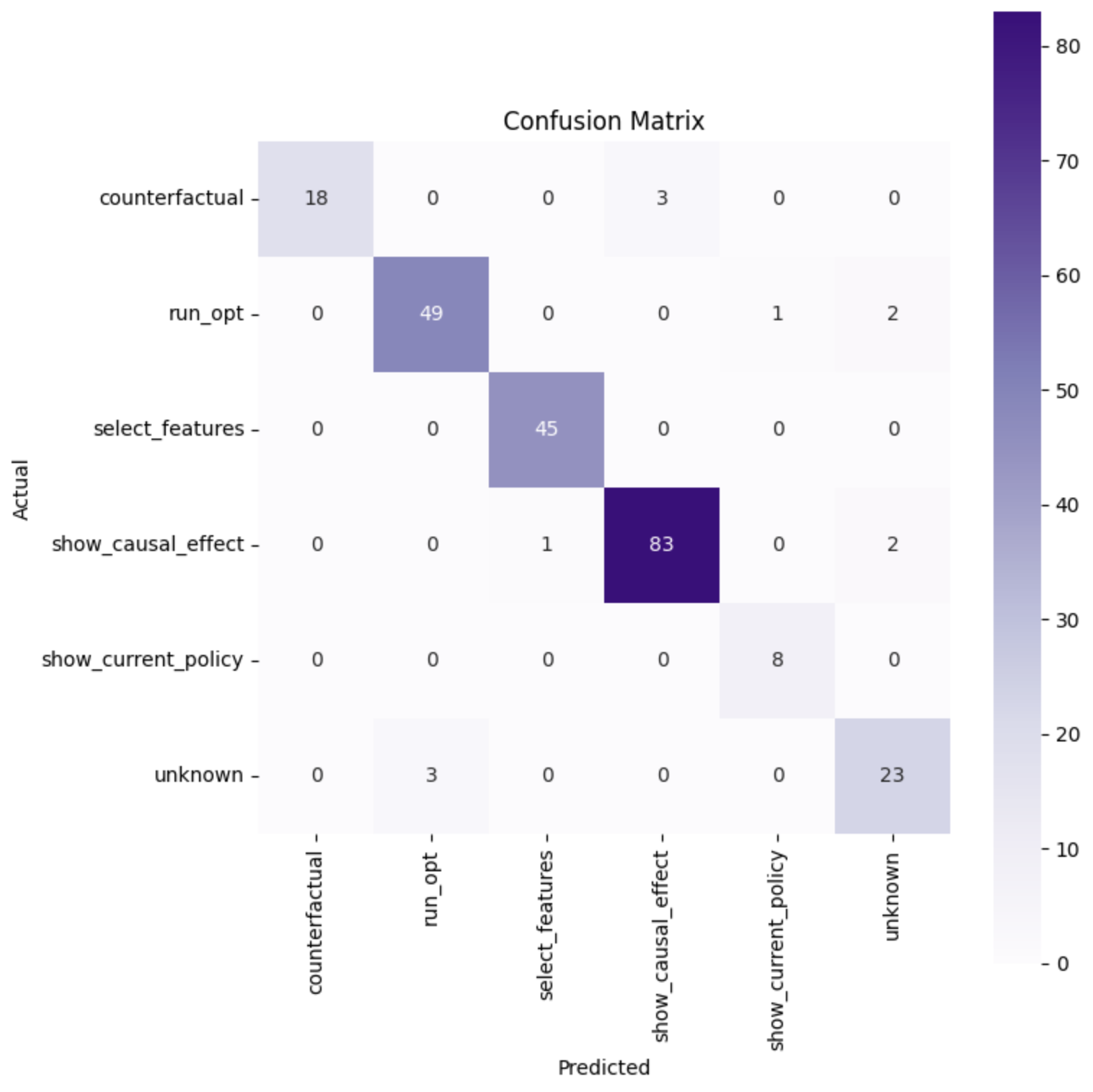}
    \caption{Confusion Matrix for Intent Classifier Prompt Tuned on Generated Training Examples}
    \label{fig:intent-confusion-matrix}
\end{figure}

In conclusion, while these experiments are not comprehensive, they do suggest that 1) the templates used to programmatically generate these training examples are capable of producing sufficiently good results when reused in other domains and 2) prompt tuning these models can potentially yield large improvements in accuracy than their ICL counterparts and lower inference times without any additional manual work - crucial for applying the agent in useful, real applications. Figure \ref{fig:intent-confusion-matrix} shows the confusion matrix for the best performing classifier.

\subsection[Evaluating Conversational Aspects]{Evaluating Conversational Aspects}\label{eval:convo}
The conversational aspects of the agent are measured qualitatively by team members to ensure that a noticeable improvement in natural responses is observed. 

\begin{figure}[htbp]
    \centering
    \begin{subfigure}[b]{1\textwidth}
        \centering
        \includegraphics[width=\textwidth]{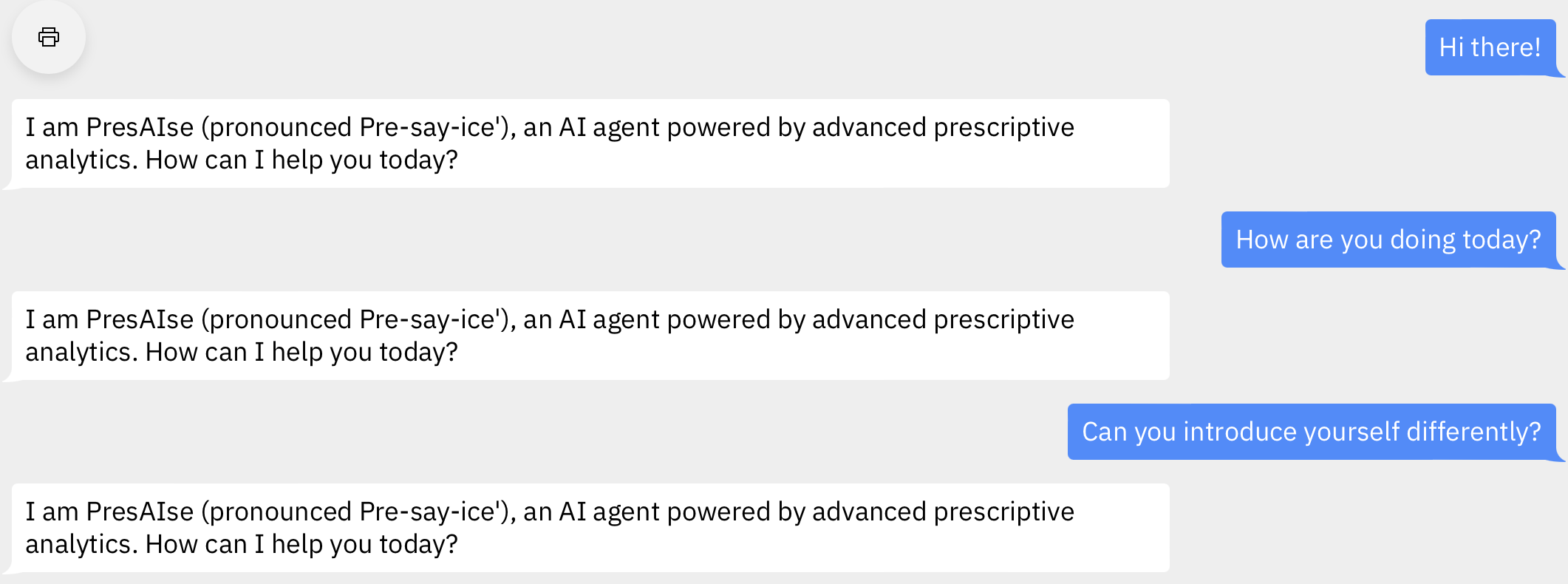}
        \caption{Few-shot prompting with FLAN model as the chat module. A rigid response is given each time.}
        \label{fig:model1}
    \end{subfigure}
    
    \begin{subfigure}[b]{1\textwidth}
        \centering
        \includegraphics[width=\textwidth]{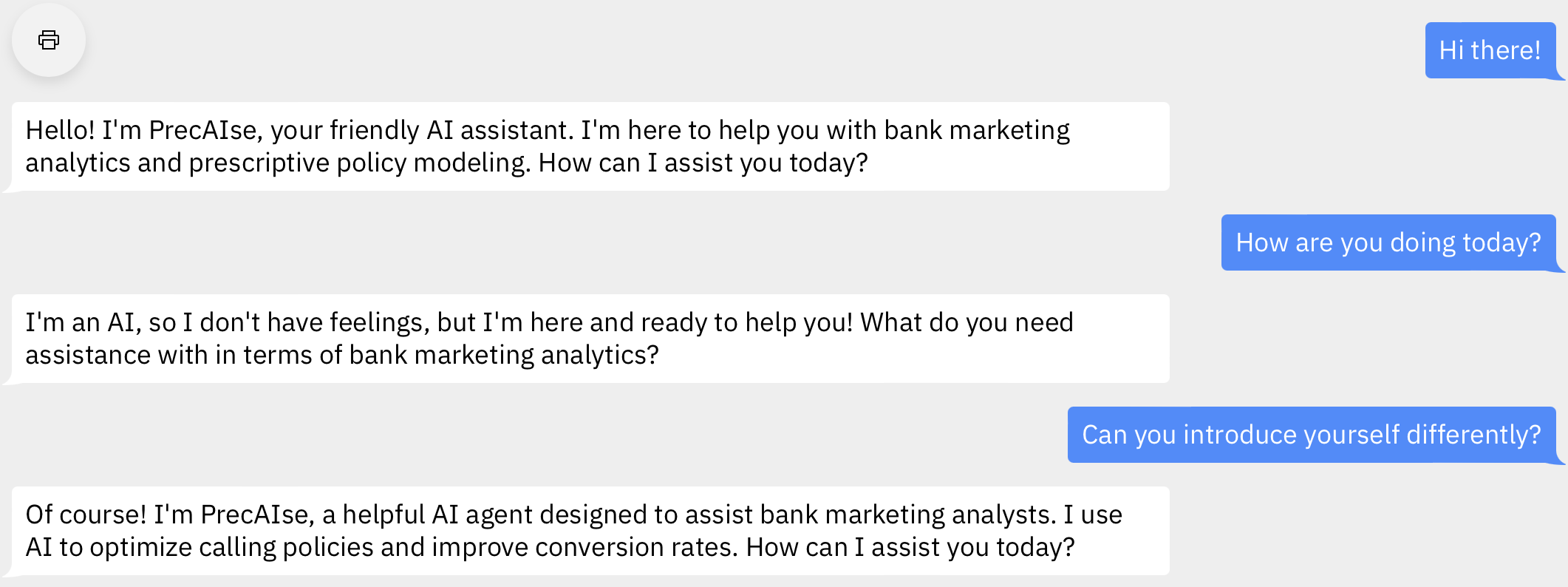}
        \caption{MoE model with larger system prompt and no few-shot examples. More diverse, natural sounding answers are given based on the query.}
        \label{fig:model2}
    \end{subfigure}
    
    \caption{Using Few-Shot Prompting with FLAN vs Using MoE Instructed to be a Chat Model}
    \label{fig:chat_model_comparison}
\end{figure}

\textbf{Chat Models.} Crafting a better system prompt as shown in Figure \ref{fig:system-prompt} and using an instruction prompted MoE model instead of relying on few-shot prompting with FLAN for the chat model produces much less "rigid" responses as seen in Figure \ref{fig:chat_model_comparison}. Notice that that chat model using ICL \ref{fig:model1} produces the same exact response to three very different queries. As expected, this response is one of the hard coded example answers seen in its few-shot examples. The chat model using MoE with a system prompt only \ref{fig:model2}, however, produces much more diverse and natural sounding answers that are more tailored to the user's query.

\begin{figure}[htbp]
    \centering
    \begin{subfigure}[b]{1\textwidth}
        \centering
        \includegraphics[width=\textwidth]{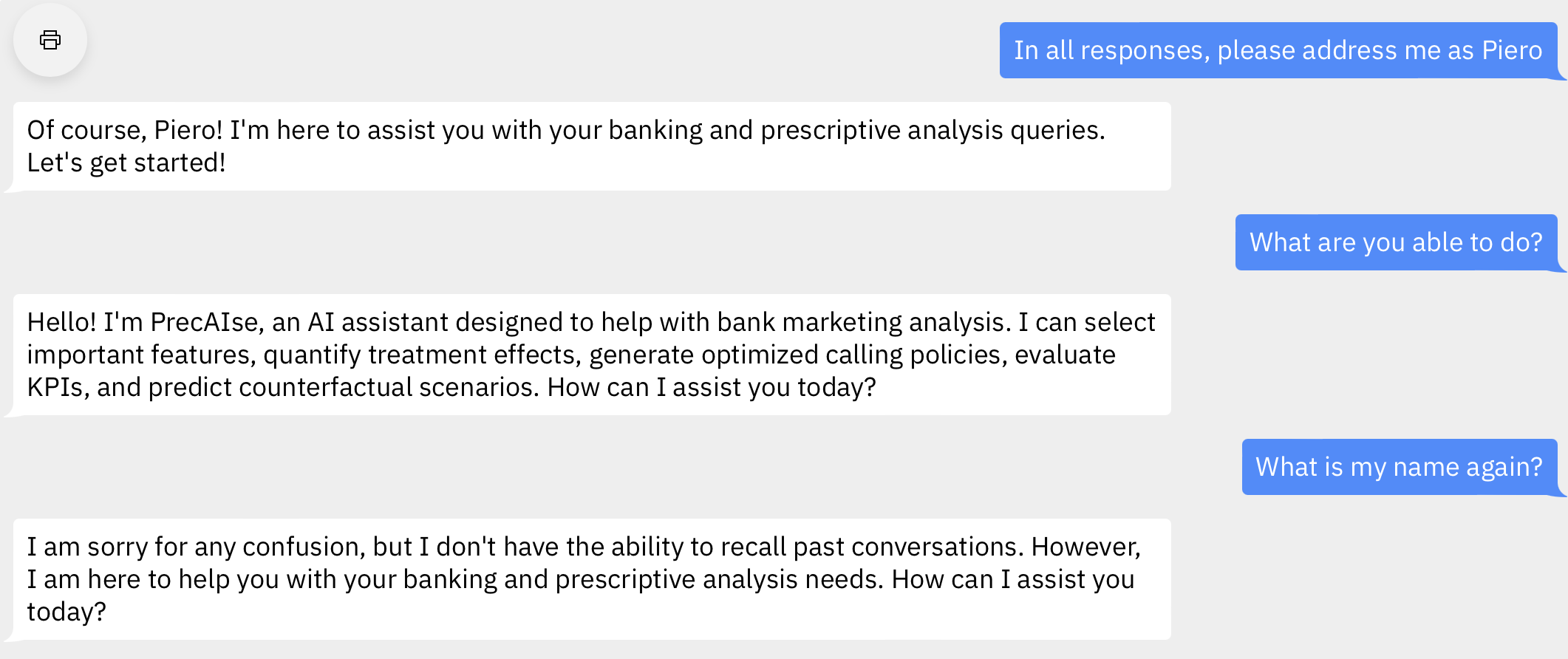}
        \caption{MoE Model without a Short Term Memory Module}
        \label{fig:mem1}
    \end{subfigure}
    
    \begin{subfigure}[b]{1\textwidth}
        \centering
        \includegraphics[width=\textwidth]{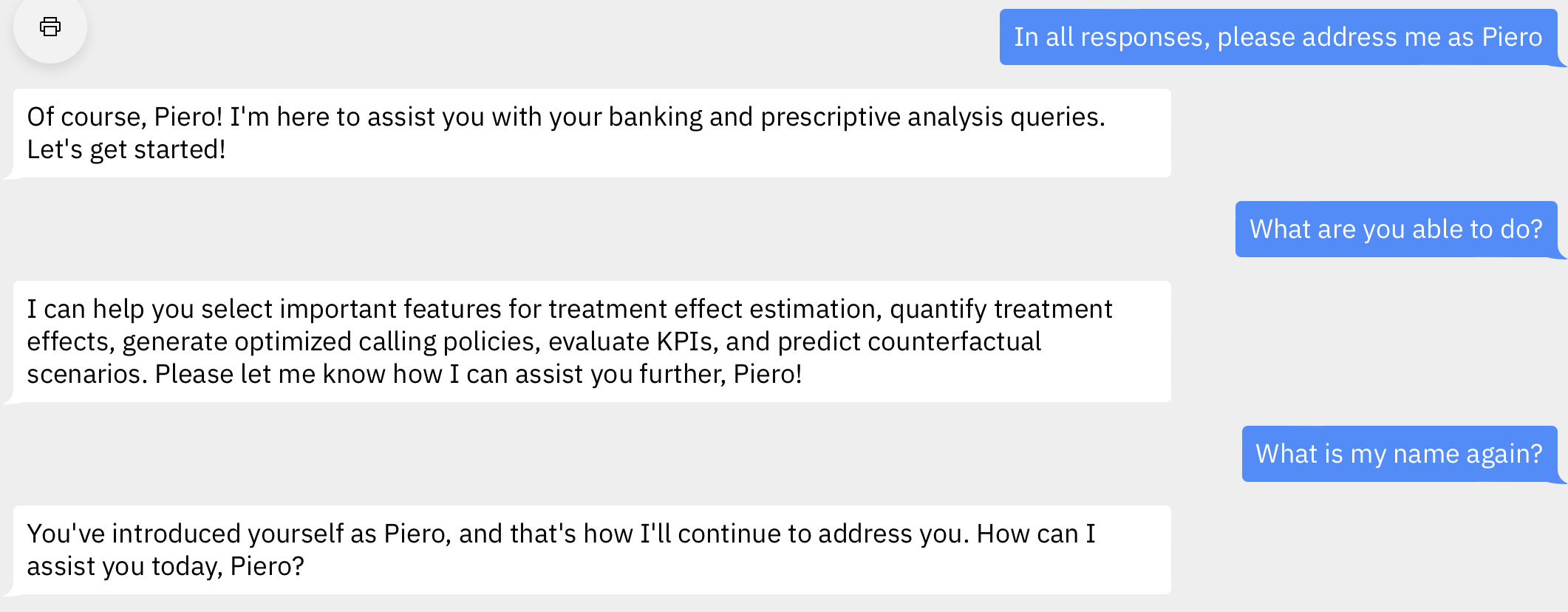}
        \caption{MoE Model with a Short Term Memory Module}
        \label{fig:mem2}
    \end{subfigure}
    
    \caption{Introducing a Short-Term Memory Module}
    \label{fig:memory_comparison}
\end{figure}

\textbf{Chat Memory.} Introducing short term memory to the agent allows for the agent to respond with context from previous questions and responses. Figure \ref{fig:memory_comparison} shows two chat modules - both using the MoE model with the system prompt in Figure \ref{fig:system-prompt} - but Figure \ref{fig:mem1} does not include the short term memory module and Figure \ref{fig:mem2} does. Notice that the model is able to remember the initial instruction and continues to reference the user by name in future responses.

\begin{figure}[htbp]
    \centering
    \begin{subfigure}[b]{1\textwidth}
        \centering
        \includegraphics[width=\textwidth]{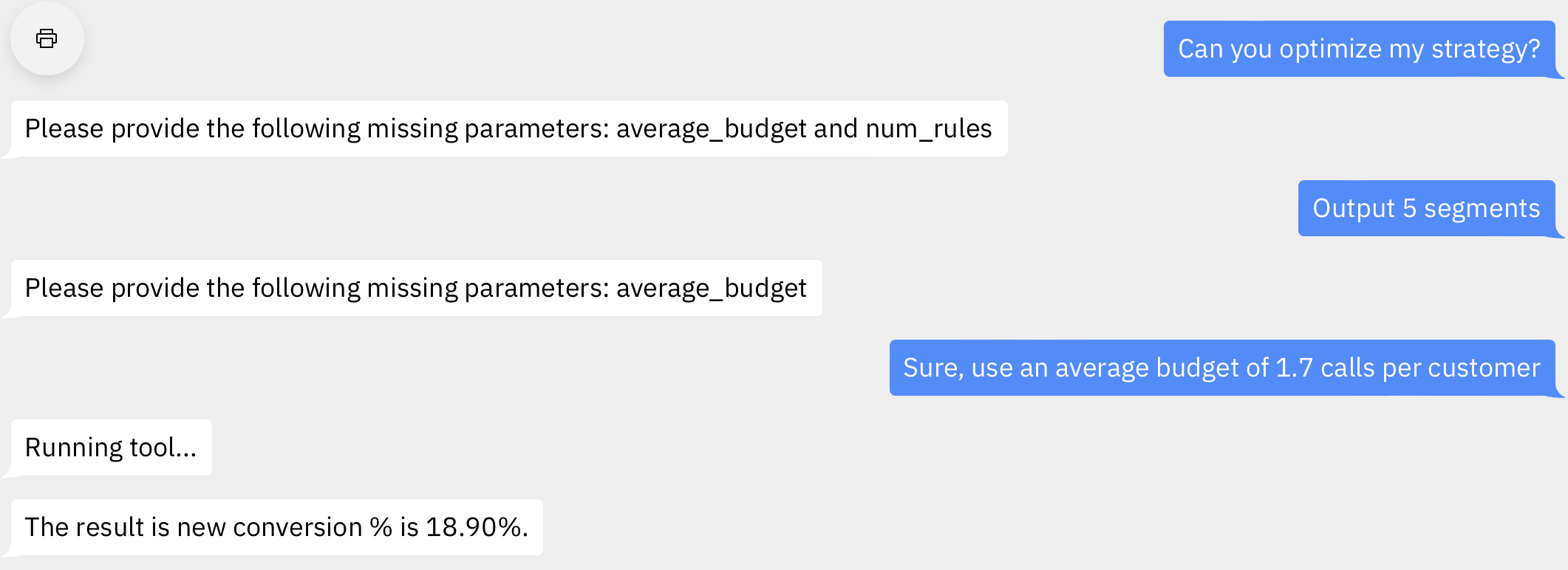}
        \caption{MoE Model without Thought Injection}
        \label{fig:nothought}
    \end{subfigure}
    
    \begin{subfigure}[b]{1\textwidth}
        \centering
        \includegraphics[width=\textwidth]{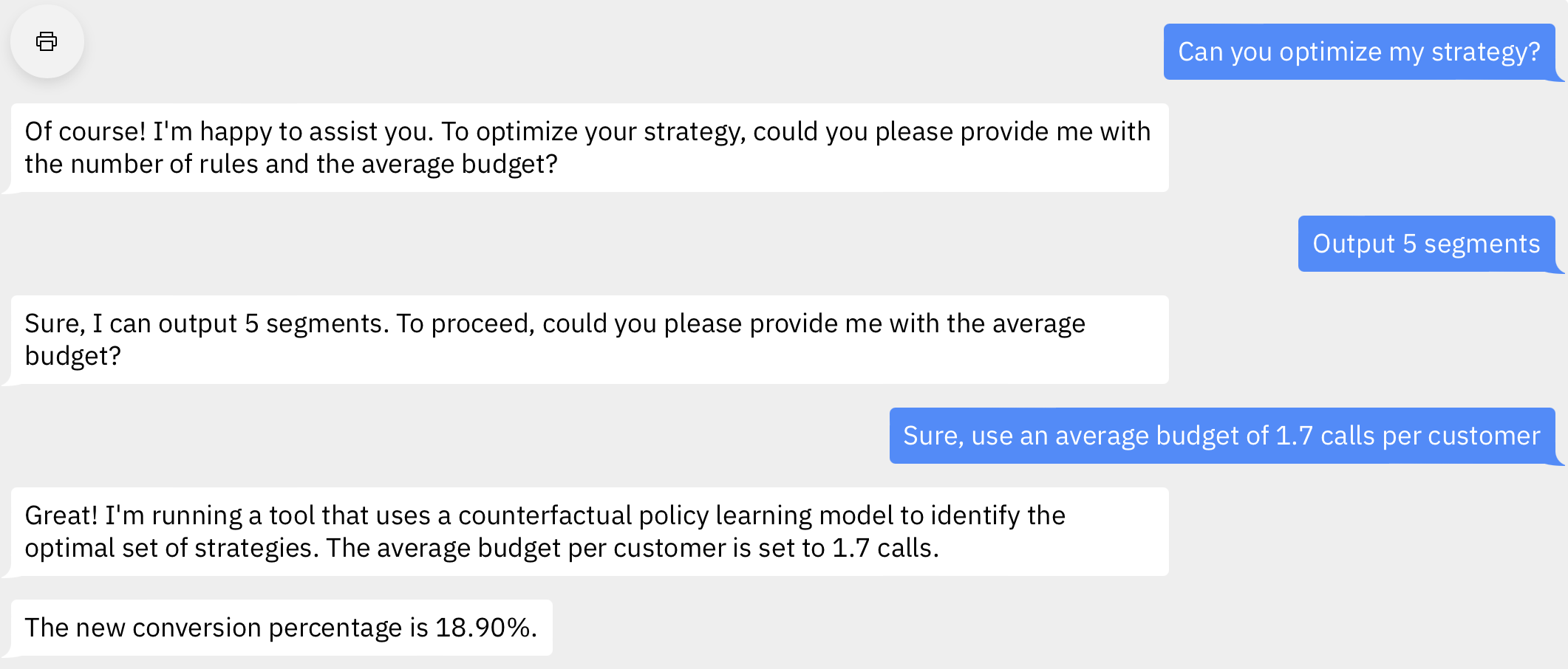}
        \caption{MoE Model with Thought Injection}
        \label{fig:withthought}
    \end{subfigure}
    
    \caption{Thought Injection Improving Natural Responses}
    \label{fig:thought_injection_hallucinate}
\end{figure}

\textbf{Hallucination.} 
The "thought injection" method is used here to help prevent false or made up information while also allowing the model to present the results instead of having a hard-coded formatted response with the results. Notice in Figure \ref{fig:thought_injection_hallucinate} the difference between having these thought injections rather than hard-coded template responses for follow-ups and results. In Figure \ref{fig:nothought}, the user is asked to provide missing parameter values in order to run the requested tool - however the follow up is very rigid, making the agent sound unnatural. Furthermore, once all the missing parameters are provided, the agent simply runs the tool and presents another templated response. Figure \ref{fig:withthought}, however, uses thought injection to instruct the MoE model to follow up and present results. It's even used to inform the user what the tool being run in the background is doing to help with the agent's credibility and interpretability.
\section[UI/UX Improvements]{UI/UX Improvements}\label{other-improvements}
User Interface (UI) and User Experience (UX) are crucial elements in the design and functionality of PrecAIse as it enhances its usability, engagement, and interpretability.

\begin{figure}[htbp]
    \centering
    \begin{subfigure}[b]{0.45\textwidth}
        \centering
        \includegraphics[width=\textwidth]{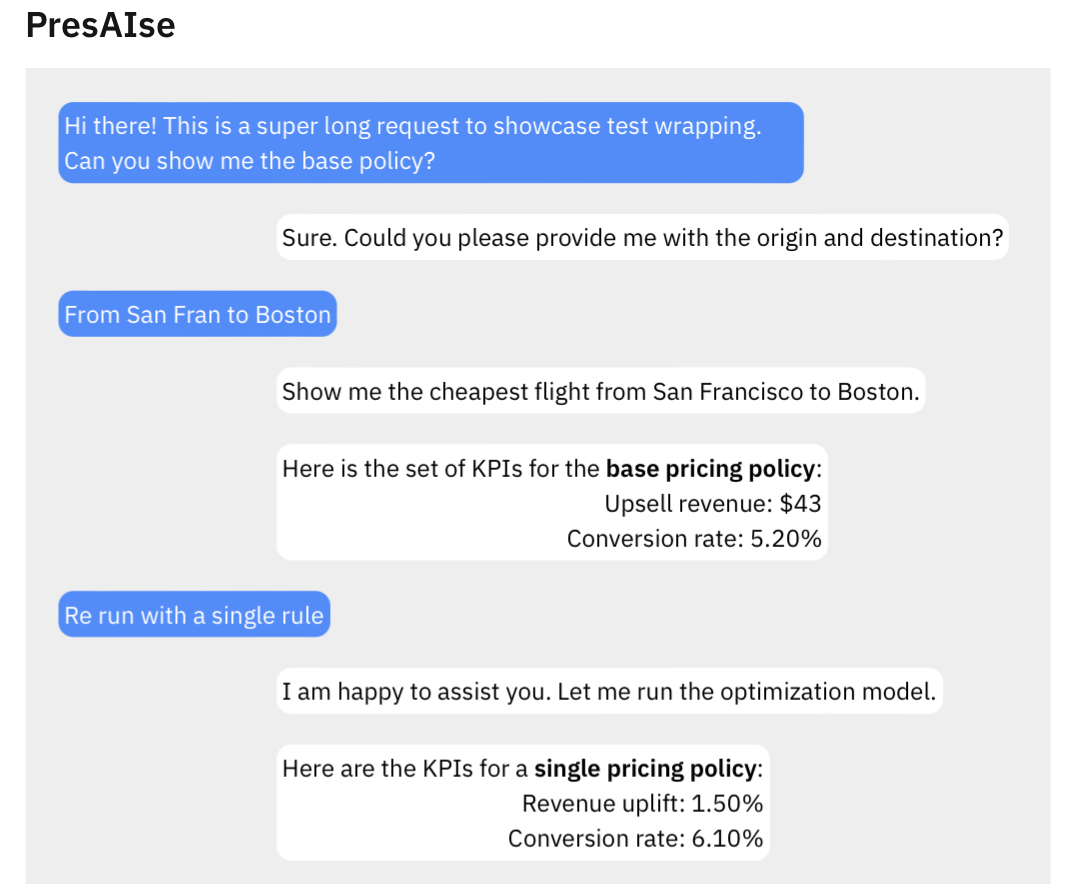}
        \caption{Original chat UI}
        \label{fig:old_chat_UI}
    \end{subfigure}
    \hspace{0.05\textwidth}
    \begin{subfigure}[b]{0.45\textwidth}
        \centering
        \includegraphics[width=\textwidth]{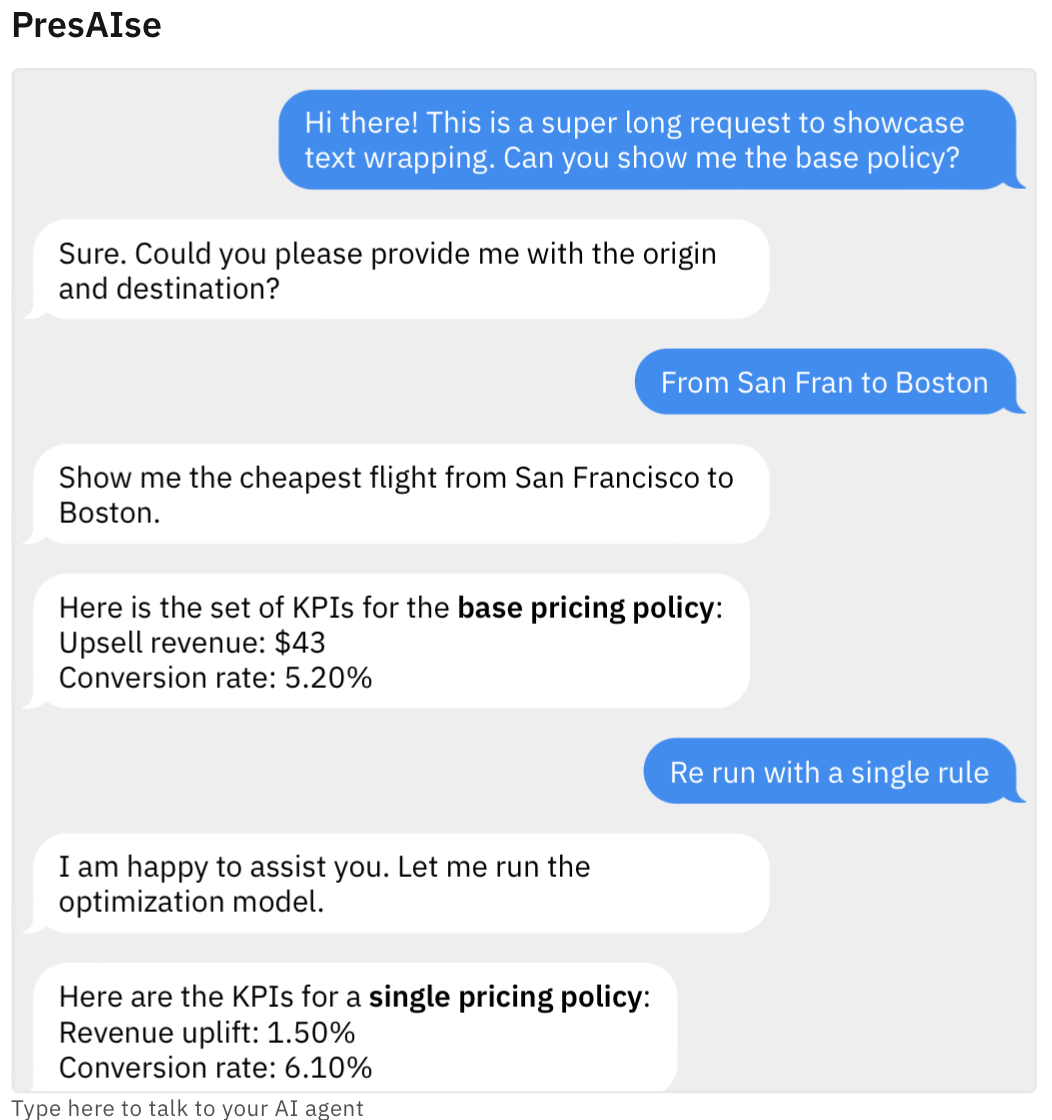}
        \caption{Updated chat UI}
        \label{fig:new_chat_UI}
    \end{subfigure}
    
    \caption{Updates to the chat UI}
    \label{fig:chat_UI}
\end{figure}

\paragraph{Chat Updates.} Changes were made to make these dialog bubbles look more "modern" and fix alignment issues that sometimes caused the agent's responses to appear in the middle of the chat window. Moreover, the order of the messages were swapped to have the user be right-aligned. This is a more common mental model used by other applications so this was done for consistency. A comparison of the original and updated UI is shown in Figure \ref{fig:chat_UI}.

\begin{figure}
    \centering
    \includegraphics[width=0.75\linewidth]{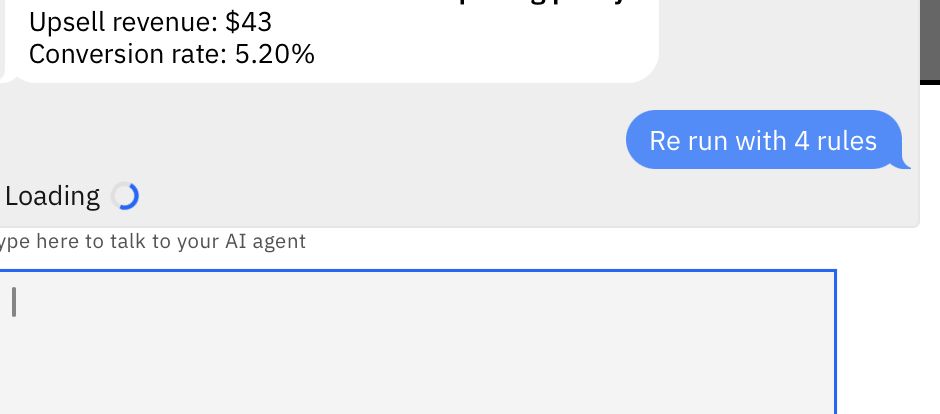}
    \caption{Loading symbol shown upon waiting for back-end response}
    \label{fig:loading}
\end{figure}

Apart from these style changes, we also added some UX changes like auto-scrolling the chat window upon new messages, having the chat input box refocus after sending a message, and adding a "loading" symbol when waiting for a response from the back-end (Figure \ref{fig:loading}).

\begin{figure}[H]
        \centering
        \includegraphics[width=1\linewidth]{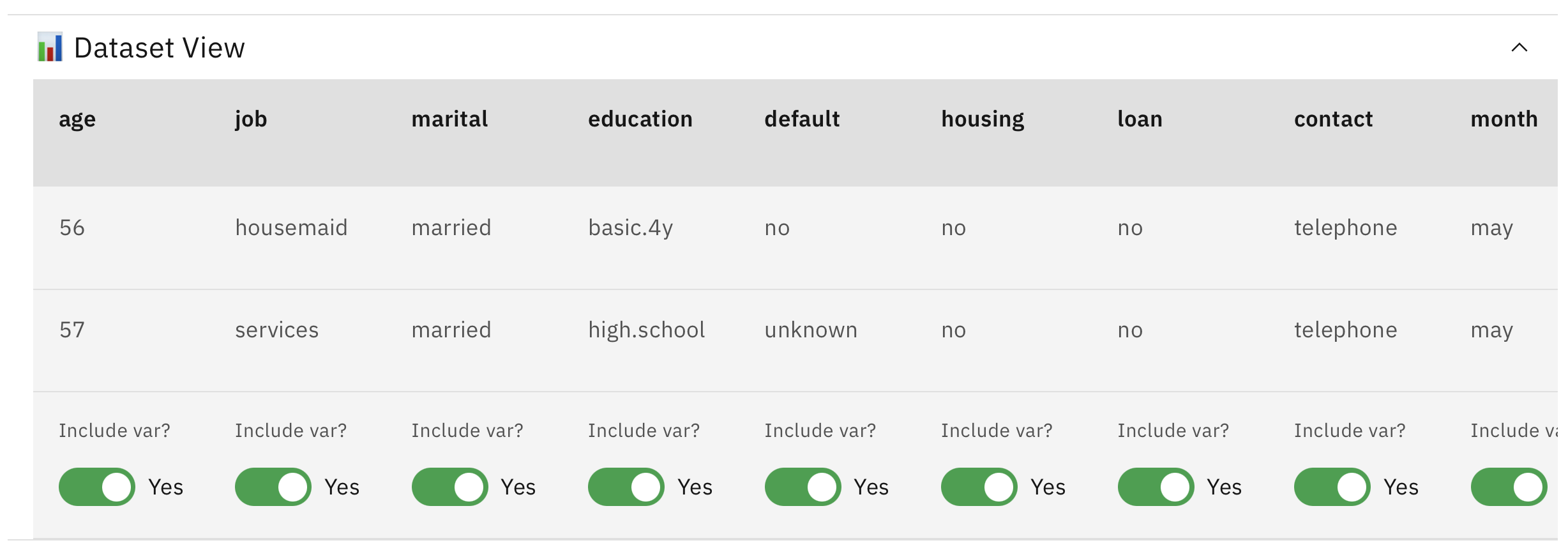}
        \caption{Dataset View}
        \label{fig:dataset_view}
\end{figure}

\paragraph{Dataset View.} Now that the agent can support new datasets, it is often helpful to have a view of the current dataset so that users know which columns they can refer to (and have the ability to "turn off" columns from the analysis). Figure \ref{fig:dataset_view} shows how this is displayed to the user.

\paragraph{Current Conditions.} With the original agent implementation, there was no way to see if the agent had understood and parsed out parameters properly. Therefore, when the agent returned results back to the user, it was difficult to know if the agent had truly used the correct function arguments without looking at the back-end logs. Additionally, there was no way to "remove" a condition or parameter once it was set. To fix these issues, we added a "Current Conditions" view that populated column and parameter values that were parsed from a user's query. This way, the user would be able to see if the agent had truly understood the request and was running the right tool with the right arguments. This helps build trust with the agent, which is important for its adoption in enterprise settings.

\begin{figure}
    \centering
    \includegraphics[width=1\textwidth]{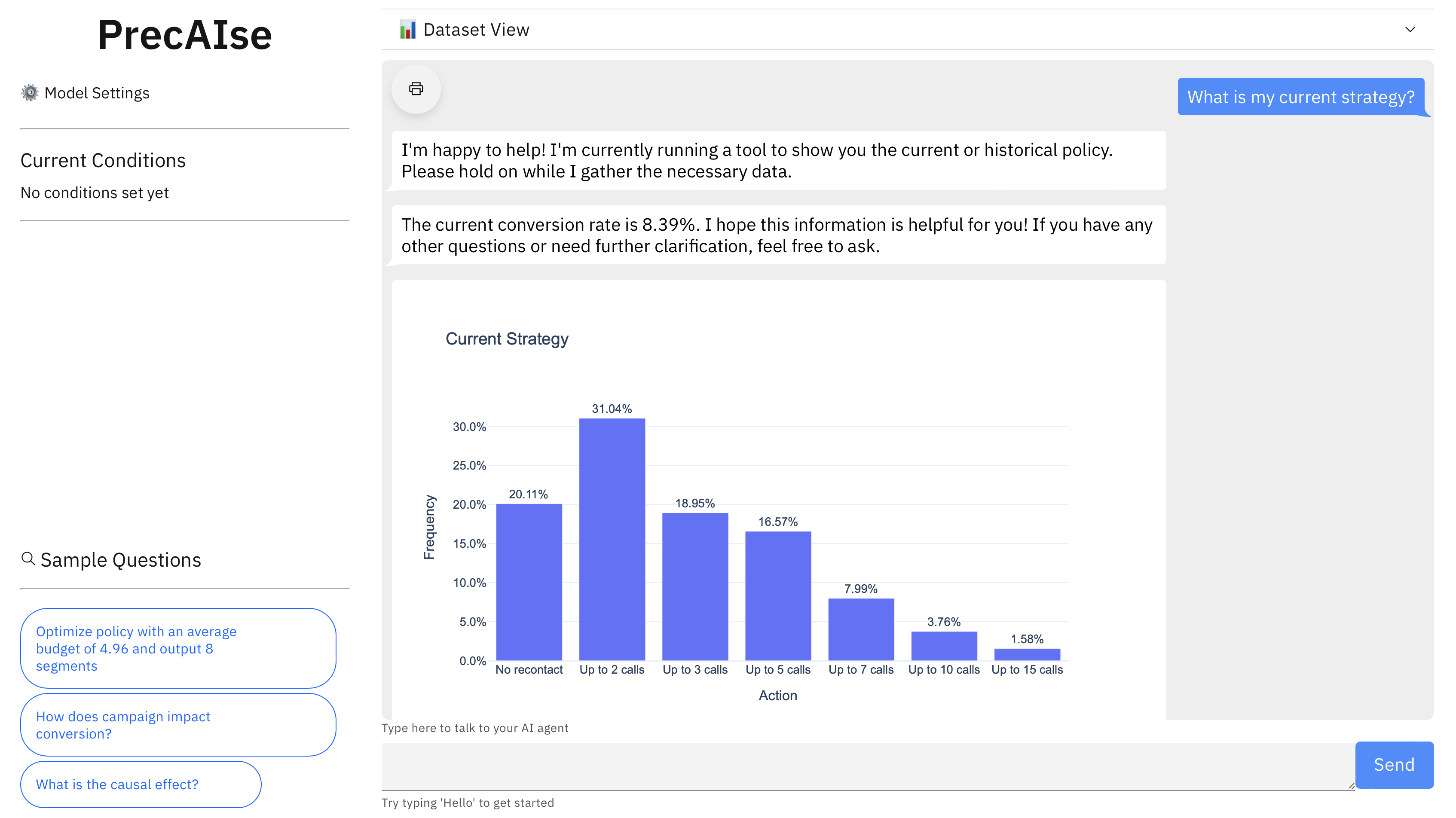}
    \caption{Final UI}
    \label{fig:final-ui}
\end{figure}

\paragraph{Sample Questions.} Another important feature for users is sample questions. We added a programmatic way to generate a couple of sample questions based on what the user has previously done. For example, if the user has just run the \textit{show\_base\_policy} tool to see what their current strategy and achieved KPI metrics are, then the sample questions might guide the user to begin exploring how their actions could affect the outcome or guide them to begin optimizing their strategy with domain-grounded parameters, as shown in Figure \ref{fig:final-ui}.

\textbf{Print Button.} Finally, we added a way for users to be able to "save" their conversations much like how data scientists can save the outputs of a Jupyter Notebook. In Figure \ref{fig:final-ui}, the print button appears at the top left-hand corner of the chat window.

\section{Related Works}

\subsection[Function Calling]{Function Calling}
Function calling in the context of LLMs refers to the ability to take a natural language query, recognize the correct function to execute, and extract out the appropriate parameters. This capability is essential for connecting LLMs with external data sources and specialized tools not included in their training data. An important consideration for enterprise use is to restrict implementations to use open-source models for cost, transparency, and flexibility reasons \cite{sun2024presAIse} to simplify integration.

\textbf{Tool Augmented Language Models.} Although the latest models from OpenAI have been trained to detect function intent and output structured JSON \cite{openai-functioncalling}, these models are proprietary and not freely available. Several studies, however, have explored integrating tools with open-source LLMs. For instance, Toolformer \cite{schick2023toolformer} is a fine-tuned GPT-J model that learns when to use tools in a self-supervised manner. Similarly, Gorilla \cite{shishir2023gorilla}, a fine-tuned LLaMA-7B model, surpasses GPT-4 in accuracy for generating API calls across extensive datasets. Despite GPT-J and LLaMA-7B being open-source, fully fine-tuning these models requires considerable computational resources, which is a limitation in the current scope.

\textbf{In-Context Learning for Intent Recognition.} Fine-tuned, tool-augmented models  (\cite{qin2023toolllm}, \cite{tang2023toolalpaca}, \cite{schick2023toolformer}, \cite{li2023apibank}, \cite{shishir2023gorilla}) demonstrate high performance, but the computational resources required for training pose a significant barrier. Providing a clear, quality instruction can be a good way to keep model parameters fixed, but even cutting-edge models like GPT-3 and GPT-4 encounter substantial difficulties in zero-shot function intent recognition \cite{li2023apibank}\cite{parikh2023intent}. This challenge makes ICL methods, such as few-shot learning, an attractive alternative, especially since it has proven effective in several tasks \cite{brown2020fewshot}. ICL potentially offers a more scalable solution than fine-tuning and can outperform zero-shot techniques for function calling.

\textbf{Parameter Efficient Fine Tuning.} One limitation of few-shot learning is its dependency on larger models \cite{brown2020fewshot}. SMASH \cite{wang2022smash} has shown promise in applying prompt tuning to smaller models, and Parikh et al. \cite{parikh2023intent} achieved notable success in intent classification using Parameter-Efficient Fine-Tuning (PEFT). PEFT methods simplify enterprise adoption, allowing businesses to use specialized models with significantly less storage space, training costs, and inference speed than other fine-tuning methods.

Low-Rank Adaptation (LoRA) \cite{liu2021lora} offers a parameter-efficient strategy  that greatly reduces the number of trainable parameters. By introducing trainable rank decomposition matrices to each transformer layer, it is able to freeze the pre-trained model weights. Although this method significantly reduces the training time, the additional weights need to be stored somewhere on enterprise servers.

\begin{figure}
    \centering
    \includegraphics[width=0.5\linewidth]{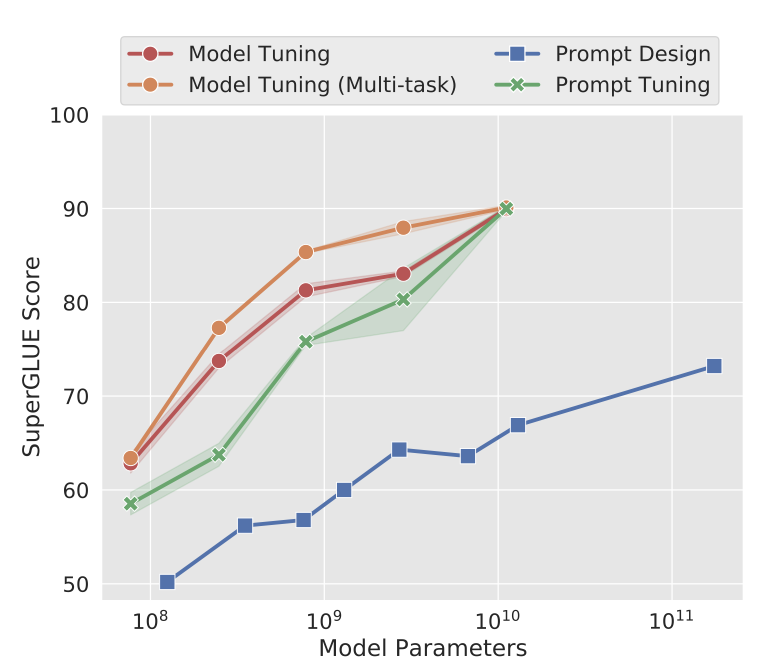}
    \caption{Prompt Tuning Performance. Adapted from \cite{lester2021powerofscale}}
    \label{fig:pt-performance}
\end{figure}

Prompt tuning presents another strategy for adapting models without modifying their weights \cite{ibm-martineau}. This approach is supported by studies showing that ``prompting often equates to hundreds of data points across classification tasks" \cite{scao2021prompt} and works by simply creating a ``learned prompt'' specific to the task. Furthermore, prompt tuning experiments show that its effectiveness approaches that of full model tuning as the number of model parameters increases \cite{lester2021powerofscale} (Figure \ref{fig:pt-performance}). While prompt tuning is more efficient than other PEFT techniques such as prefix-tuning, the optimized prompt is in general not interpretable \cite{ibm-martineau}.

\subsection[Conversational Aspects]{Conversational Aspects}
Functions that make up a usable and interpretable agent includes the capability of engaging in natural conversations, providing appropriate follow-ups, and ensuring that the information it presents or requests does not stem from hallucinated inputs, outputs, or data.

\textbf{Models.} Different models have been fine-tuned for specific tasks or task groups. Google's Fine-tuned Language Net (FLAN) \cite{chung2022googleflan} has been instruction-tuned on various tasks like question and answering and various chain-of-thought tasks. Because of this, the model demonstrates considerable proficiency in handling tasks it was not directly trained on. This means that it is possible to use instruction-tuned models as chat models by simply instructing the model to act like a chat model. However, dedicated chat models, which are specifically trained to manage conversational dynamics, could potentially offer improved performance. Mixture of Experts (MoE) models can also be a superior alternative due to its faster inference times compared to similarly sized models \cite{sanseviero2023MoE}.

\textbf{Chat Memory.} Effective memory utilization is crucial for agents to maintain an enhanced contextual understanding and produce responses that appear more natural to users. Different types of memory include short-term memory (for recent interactions), long-term memory, episodic memory (for significant interactions), and semantic memory (for general knowledge not tied to specific instances) \cite{kashyap2023memory}. Selecting the appropriate type of memory is vital, as each has its own set of advantages in regards to complexity, scalability, privacy, and inference time.

\textbf{Hallucination.} Hallucination occurs when a model produces a response that seems plausible but in actuality is unfaithful \cite{Ji2023Hallucination}. Important for decision-making, responses must be anchored in factual data. While techniques like Retrieval Augmented Generation (RAG) \cite{rag} and Knowledge Retrieval \cite{knowledge_retrieval} have been used to mitigate hallucinations, these techniques often introduce latency and maintenance overhead. Employing chain-of-thought prompting has been shown to enhance the accuracy of outputs from LLMs \cite{wei2022CoT}, and prompt injection - which is "the process of overriding original instructions in the prompt with special user input" \cite{prompt_injecttion} - can strongly align model responses.
\section[Conclusion]{Conclusion}\label{conclusion}
The culmination of this work is presented in a final demonstration of the agent in Appendix \ref{A:demonstration}. While results with the prompt tuned models for intent recognition and parameter extraction to support function calling were not exhaustive, they did serve to give further insights into the capabilities of these smaller models (Google's FLAN T5 XL model has 3B parameters \cite{chung2022googleflan}). Furthermore, is has been shown that the agent architecture presented is capable of working well and adapting to new domains, such as customer churn datasets. Techniques like ``thought injection'' were presented to help with alignment and hallucinations - proving that these issues can be mediated in use cases that require a professional and accurate agent.
Further improvements to the agent can be made by focusing on supporting more tools. While it is suspected that prompt tuning may handle the intent classification of many more tools, there are potentially scaling issues with creating parameter extractors for each possible parameter among these tools. Furthermore, for use cases with strict resource limits, it is worth exploring how multiple tools may be supported without the use of any fine-tuning at all.

\section*{Acknowledgements} 
We would like to express our deepest appreciation to Scott McFaddin at IBM Research for his support and invaluable feedback during this project. We also extend our gratitude to other members at the MIT-IBM Watson AI Lab, including Onkar Bhardwaj, Veronique Demers, and Mikhail Yurochkin, for their valuable insights and advocacy. Finally, we are grateful to Piero's thesis advisor, Devavrat Shah, for his thorough reviews and guidance.


\appendix

\section{Demonstration}\label{A:demonstration}

\lstdefinestyle{mystyle}{
    backgroundcolor=\color{CadetBlue!15!white},   
    commentstyle=\color{Red3},
    numberstyle=\tiny\color{gray},
    stringstyle=\color{Blue3},
    basicstyle=\small\ttfamily,
    breakatwhitespace=false,         
    breaklines=true,                 
    numbers=left,                    
    numbersep=5pt,                  
    showspaces=false,                
    showstringspaces=false,
    showtabs=false,                  
    tabsize=2
}%
\lstset{language=[5.3]Lua,style={mystyle}}%

\begin{figure}[H]
    \centering
    \includegraphics[width=1\linewidth]{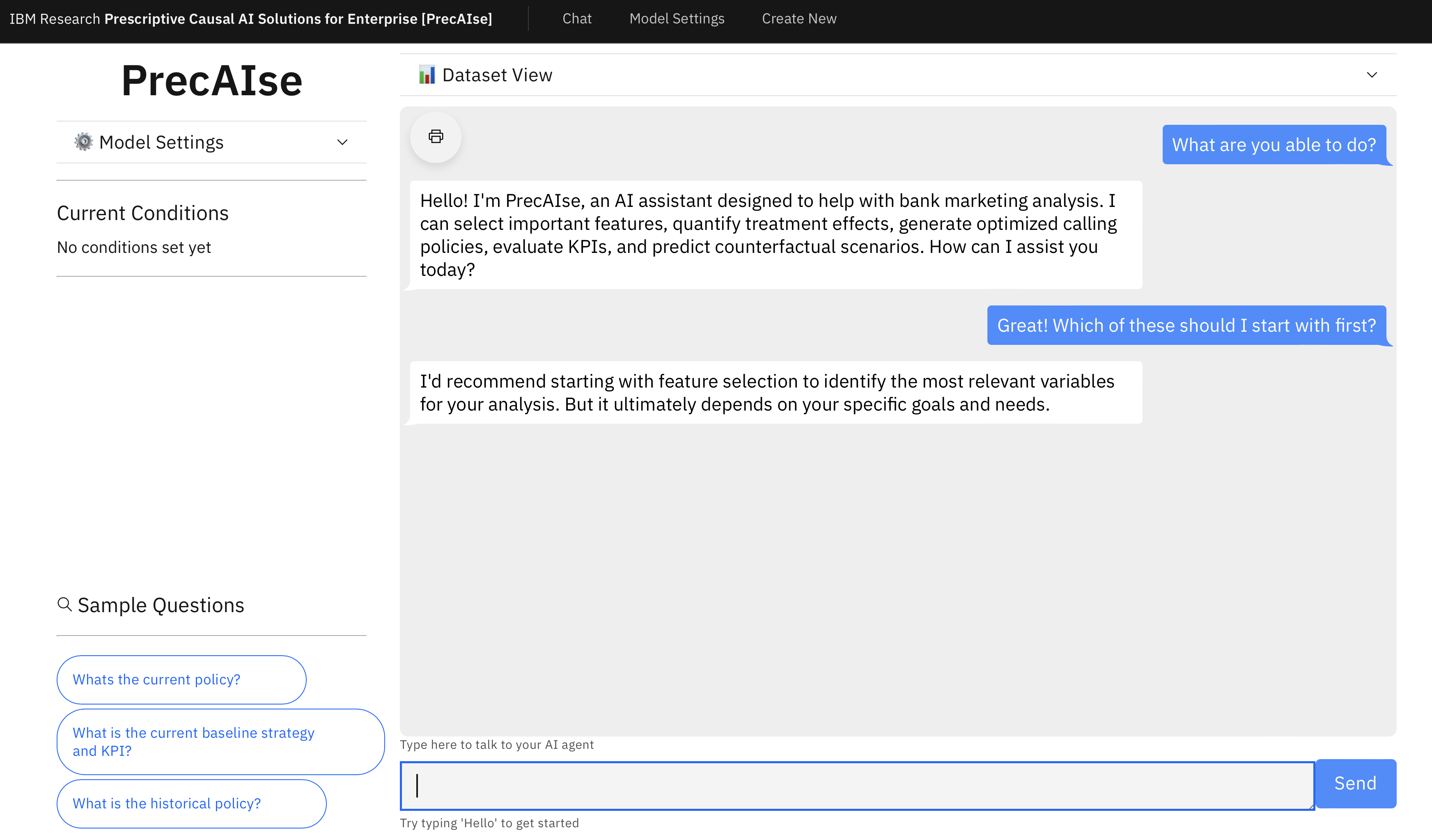}
    \caption{Initial exploration}
    \label{fig:dem-landing}
\end{figure}

This section shows a demonstration of the final deliverable. This use case is for an open source bank marketing dataset \cite{bank_marketing_222}. We assume the role of a telemarketing agent who is trying to optimize their campaign strategy, or the number of times to call a client, to maximize the number of clients they can get to subscribe to a fixed term deposit, constrained by an average budget per client. So in this scenario, the "action" variable is the "CAMPAIGN" column, and the outcome variable corresponds to the "CONVERSION" column. We start with Figure \ref{fig:dem-landing}, asking the agent about its capabilities, much like a telemarketer user would do. Thanks to the conversational LLM's new semantic memory added in Section \ref{sec:convo_methods}, the agent is even able to recommend one of the actions it previously outputted.

\begin{figure}[H]
    \centering
    \includegraphics[width=0.8\linewidth]{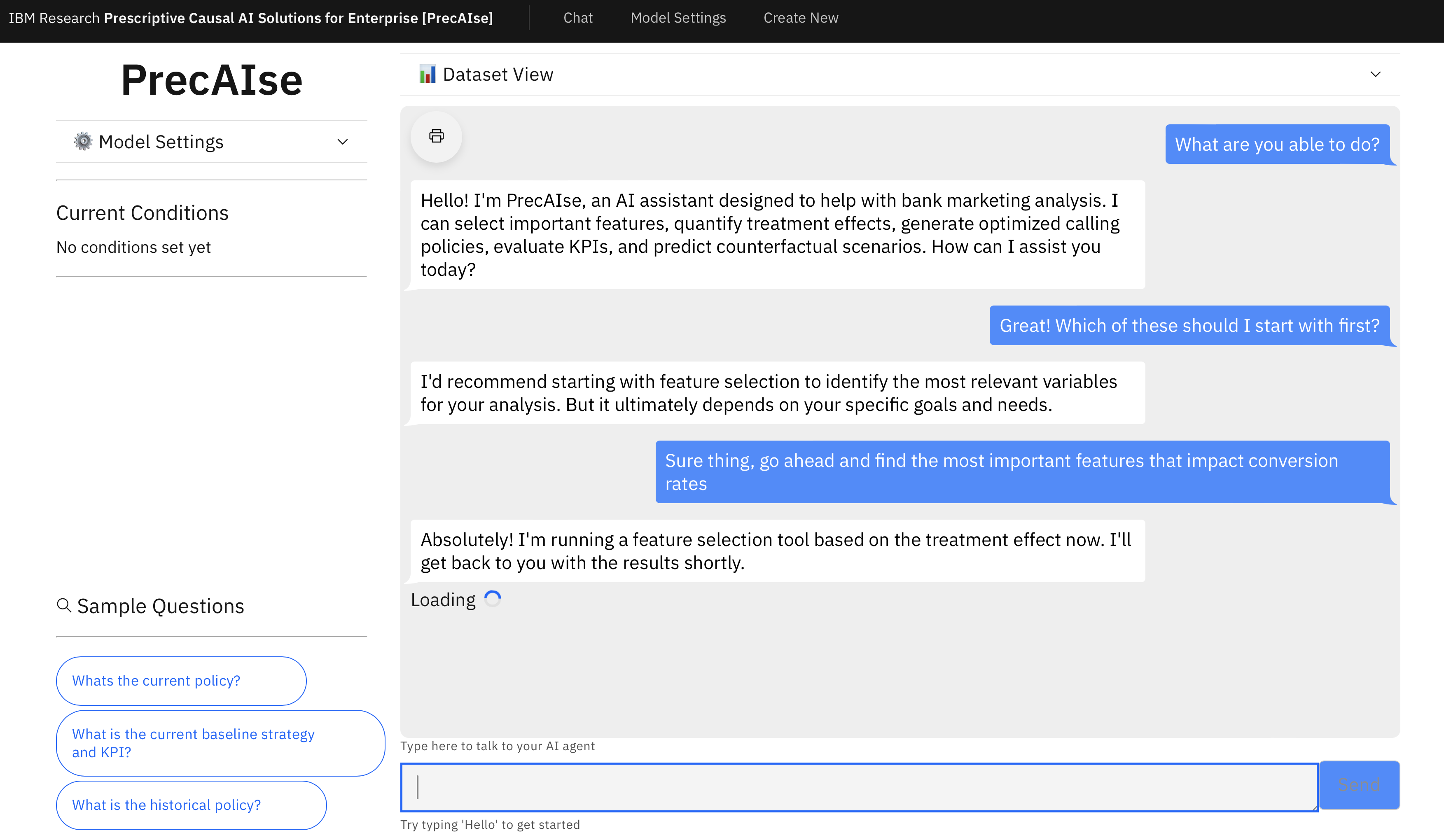}
    \caption{Agent responds immediately while tool executes}
    \label{fig:feature_selection}
\end{figure}

We accept the agent's recommendation and ask it to identify the most important features. Because this \textit{select\_features} tool can take up to 2 minutes to execute, the agent returns an immediate response letting the user know what is being done in background in Figure \ref{fig:feature_selection}. 

\begin{figure}[H]
    \centering
    \includegraphics[width=0.8\linewidth]{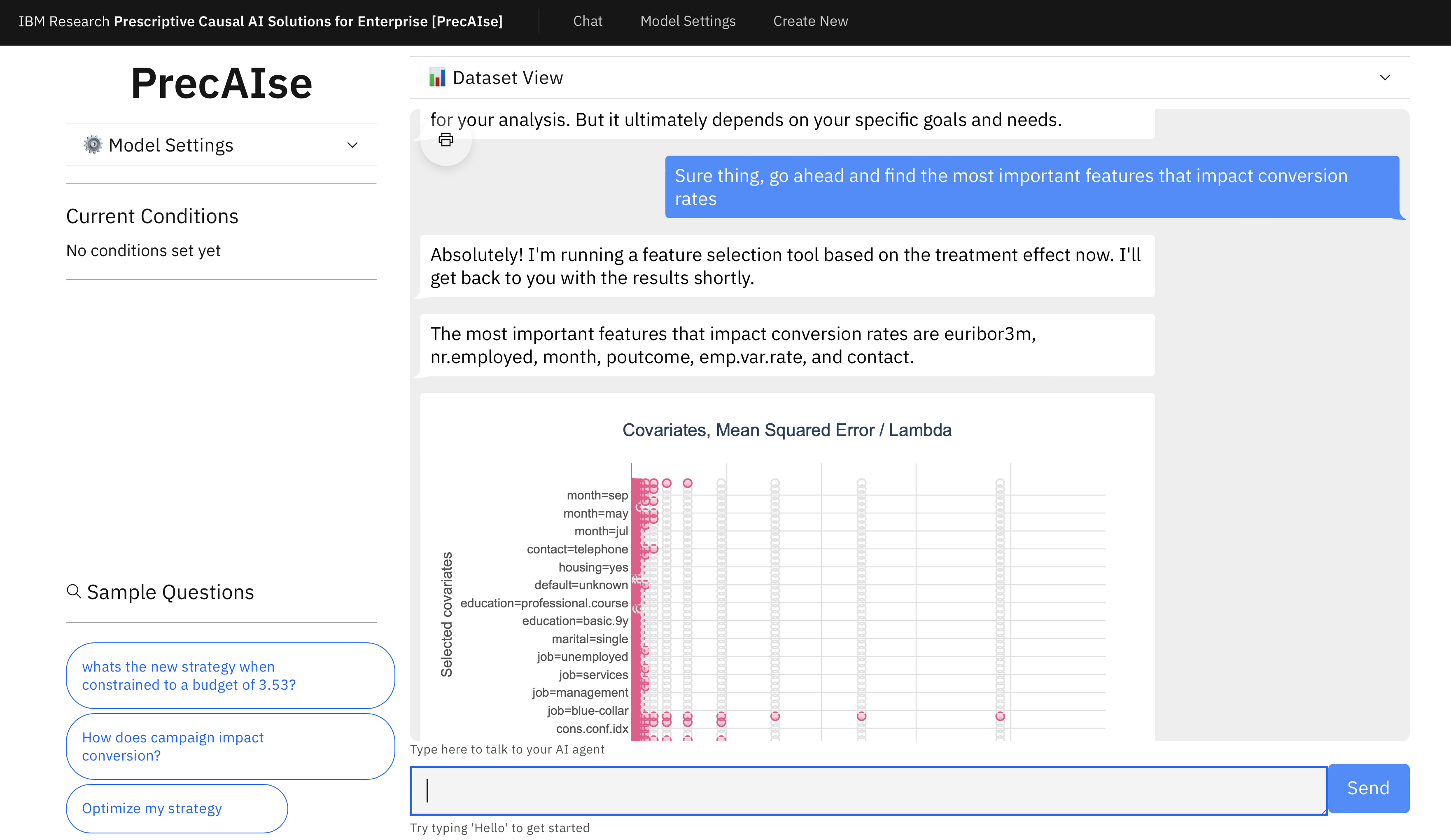}
    \caption{Feature selection tool}
    \label{fig:feature-result}
\end{figure}

Once the tool is finished executing, the agent presents and formatted numerical/textual data and any plots returned by the tool as shown in Figure \ref{fig:feature-result}.

\begin{figure}
    \centering
    \includegraphics[width=0.8\linewidth]{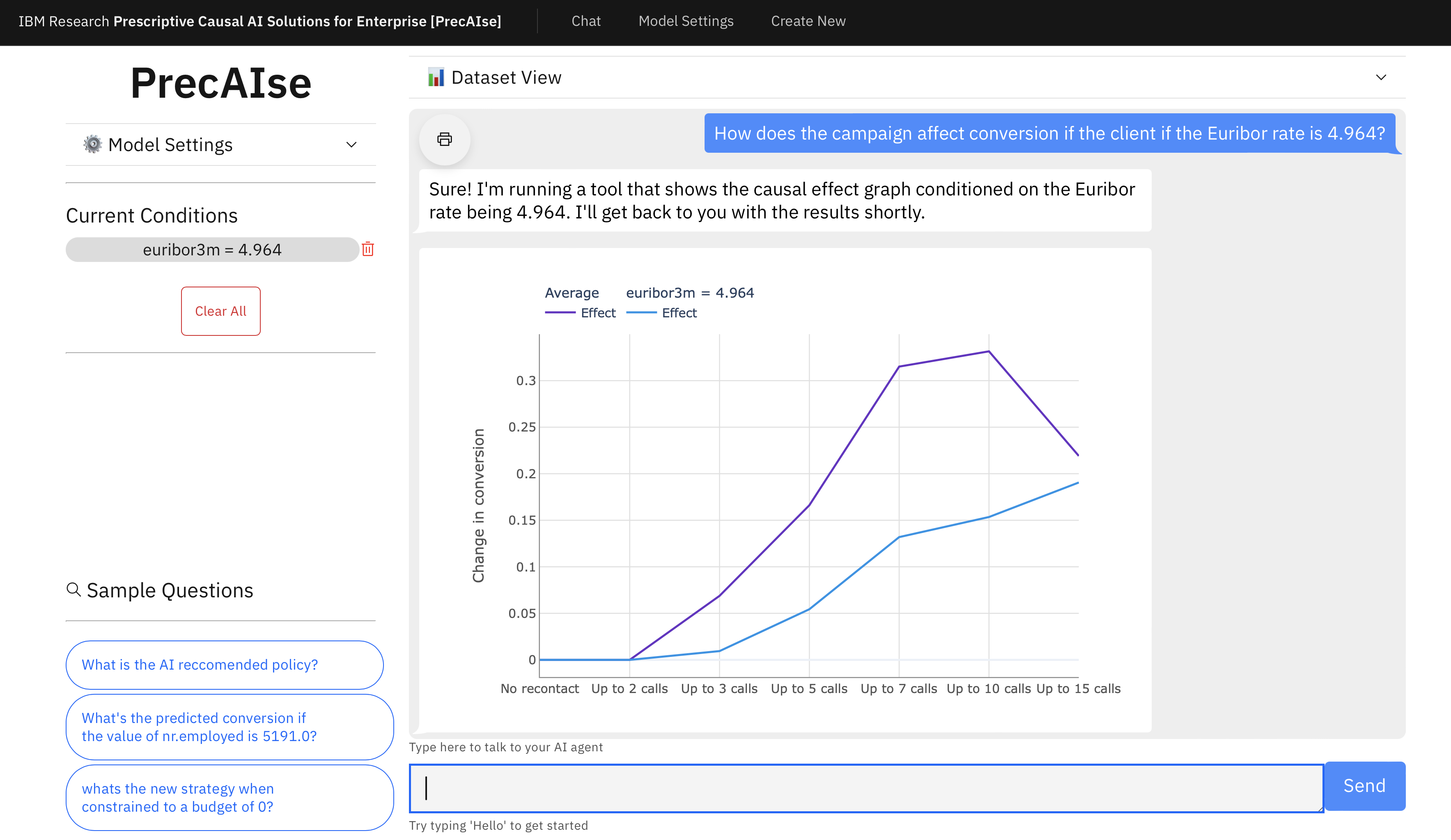}
    \caption{Causal effect tool}
    \label{fig:causal_effect_tool}
\end{figure}

Next, the telemarketer might want to do further exploratory analysis - asking questions about how campaigning affects conversion conditioned on some features having set values. In Figure \ref{fig:causal_effect_tool}, we ask the agent how the chances of conversion is impacted by our calling strategy when we condition on the Euro Interbank Offered Rate, or Euribor, being equal to $4.964$ - triggering the \textit{counterfactual} tool in the background. Notice that 1) the side bar has been populated with this new condition and 2) the returned plot shows how the chances of conversion differ from the average and the conditioned case.

\begin{figure}[H]
    \centering
    \includegraphics[width=0.8\linewidth]{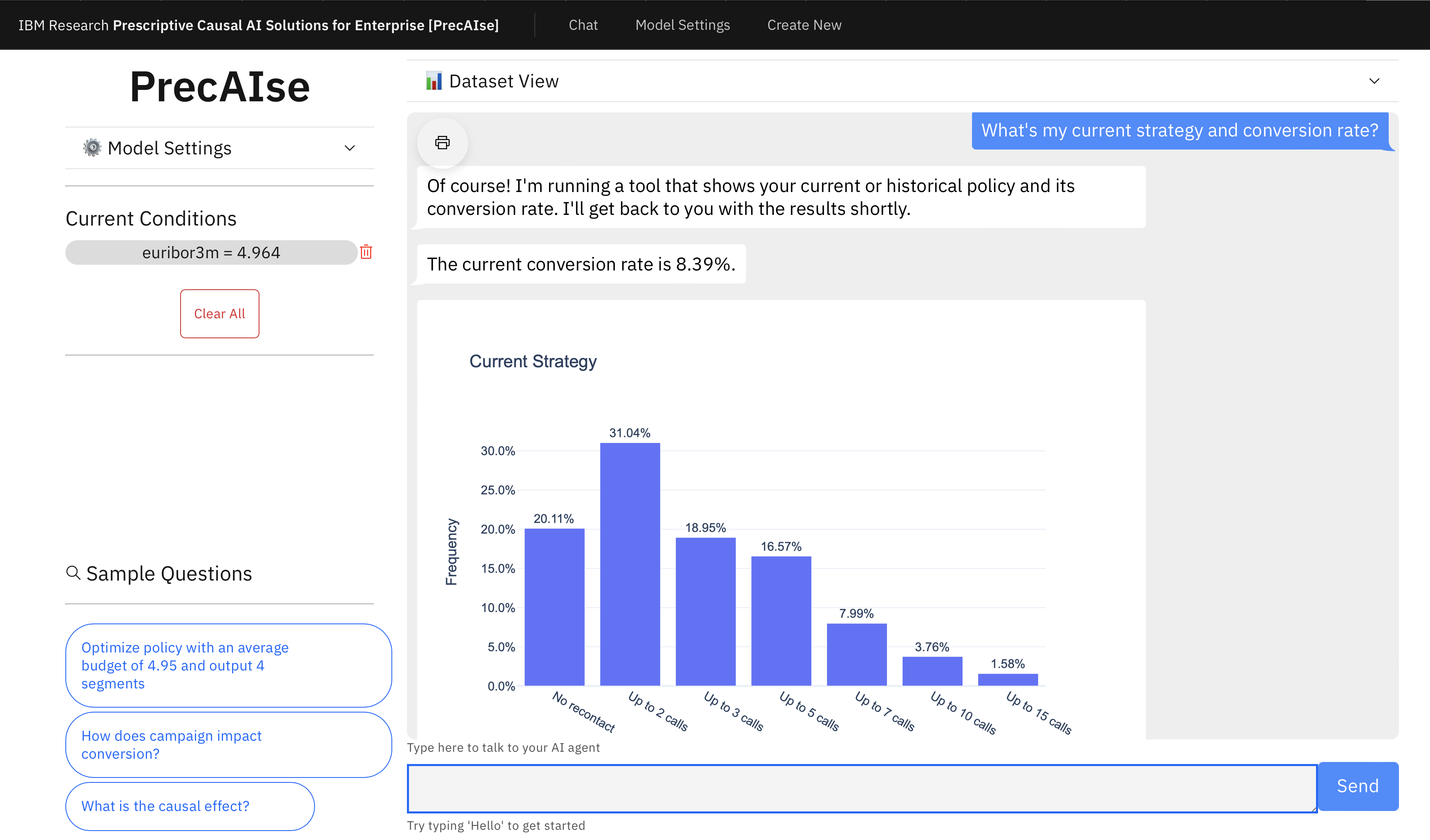}
    \caption{Current strategy tool}
    \label{fig:curr_strat}
\end{figure}

Next, the telemarketer might want to know what their current campaign strategy is, and the conversion rate that they are achieving. Executing the \textit{show\_base\_policy} tool, Figure \ref{fig:curr_strat} shows that most clients are either receiving up to 2 reminder calls or none at all, and that this strategy is currently producing a conversion rate of $8.39\%$.

\begin{figure}[H]
    \centering
    \includegraphics[width=0.8\linewidth]{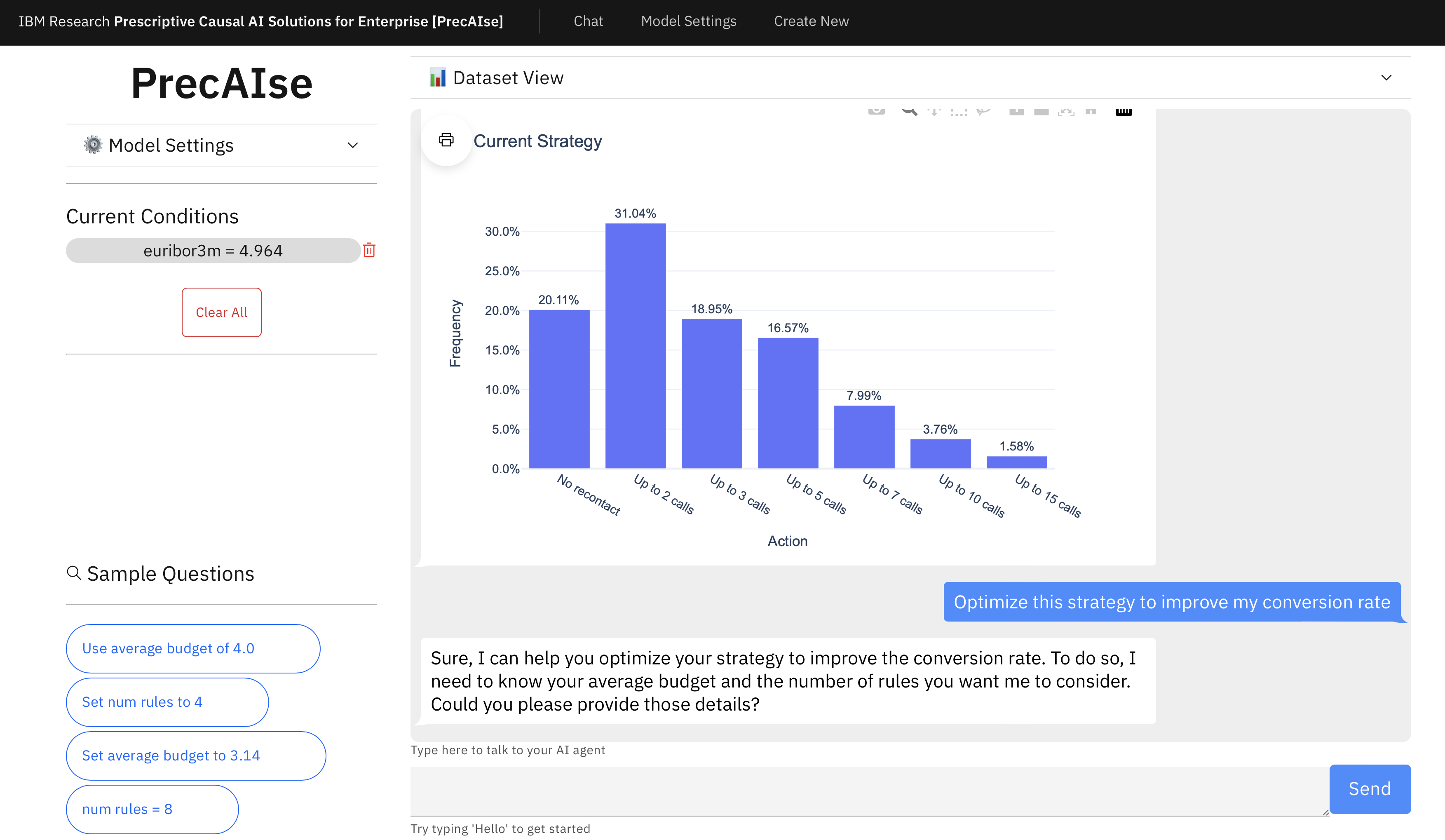}
    \caption{Follow-up for missing parameters}
    \label{fig:followup1}
\end{figure}

Naturally from here, the telemarketer may want the agent to optimize this strategy in order to improve the conversion rate. Notice in Figure \ref{fig:followup1}, the agent recognizes that the \textit{run\_optimize} tool requires two missing parameters, so it follows up with user.

\begin{figure}[H]
    \centering
    \includegraphics[width=0.8\linewidth]{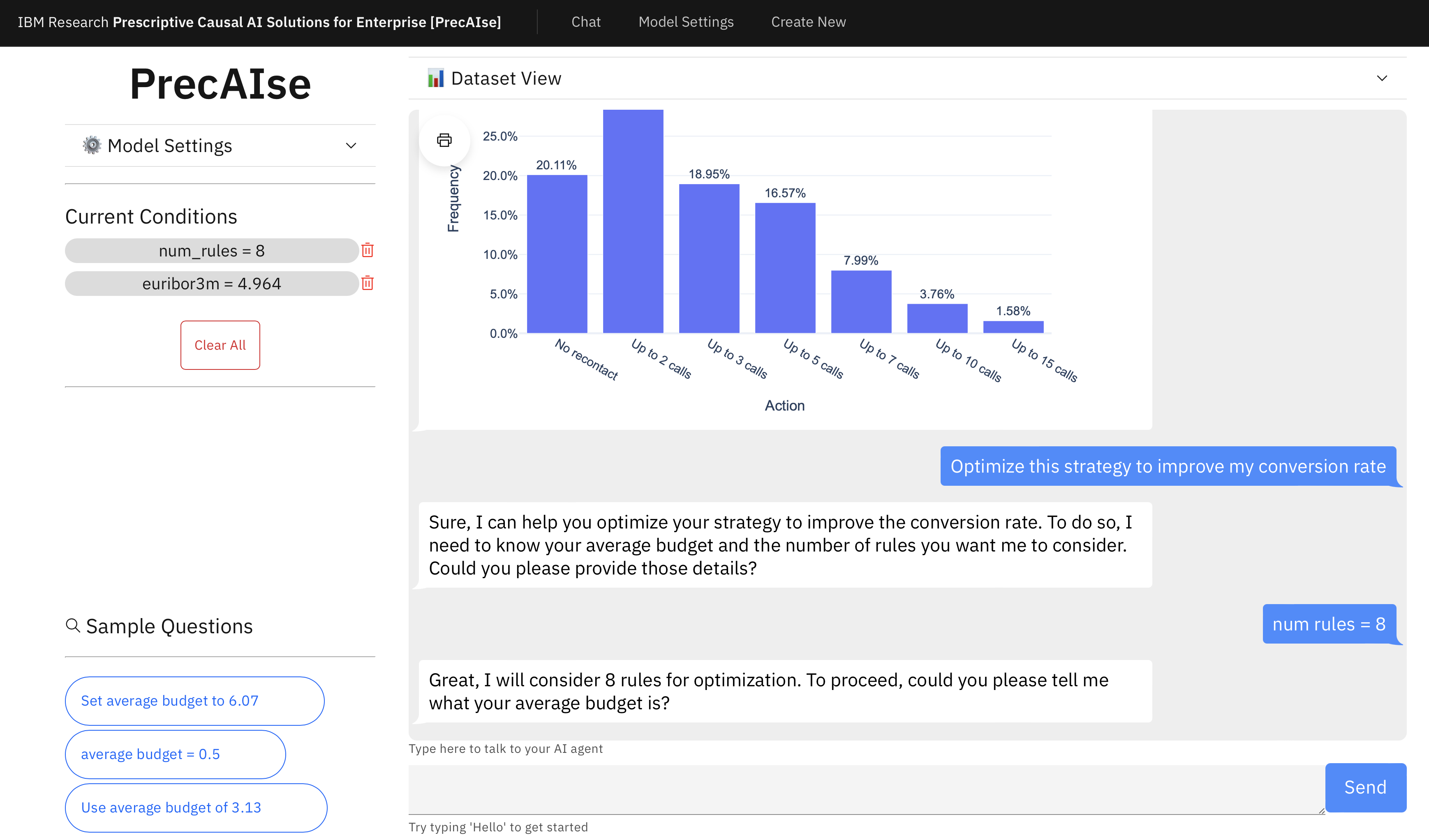}
    \caption{Follow-up for missing budget}
    \label{fig:followup2}
\end{figure}

Using the sample questions, we provide the "num rules" parameter. The agent follows up again in Figure \ref{fig:followup2}, requesting that the average budget parameter be supplied.

\begin{figure}[H]
    \centering
    \includegraphics[width=0.8\linewidth]{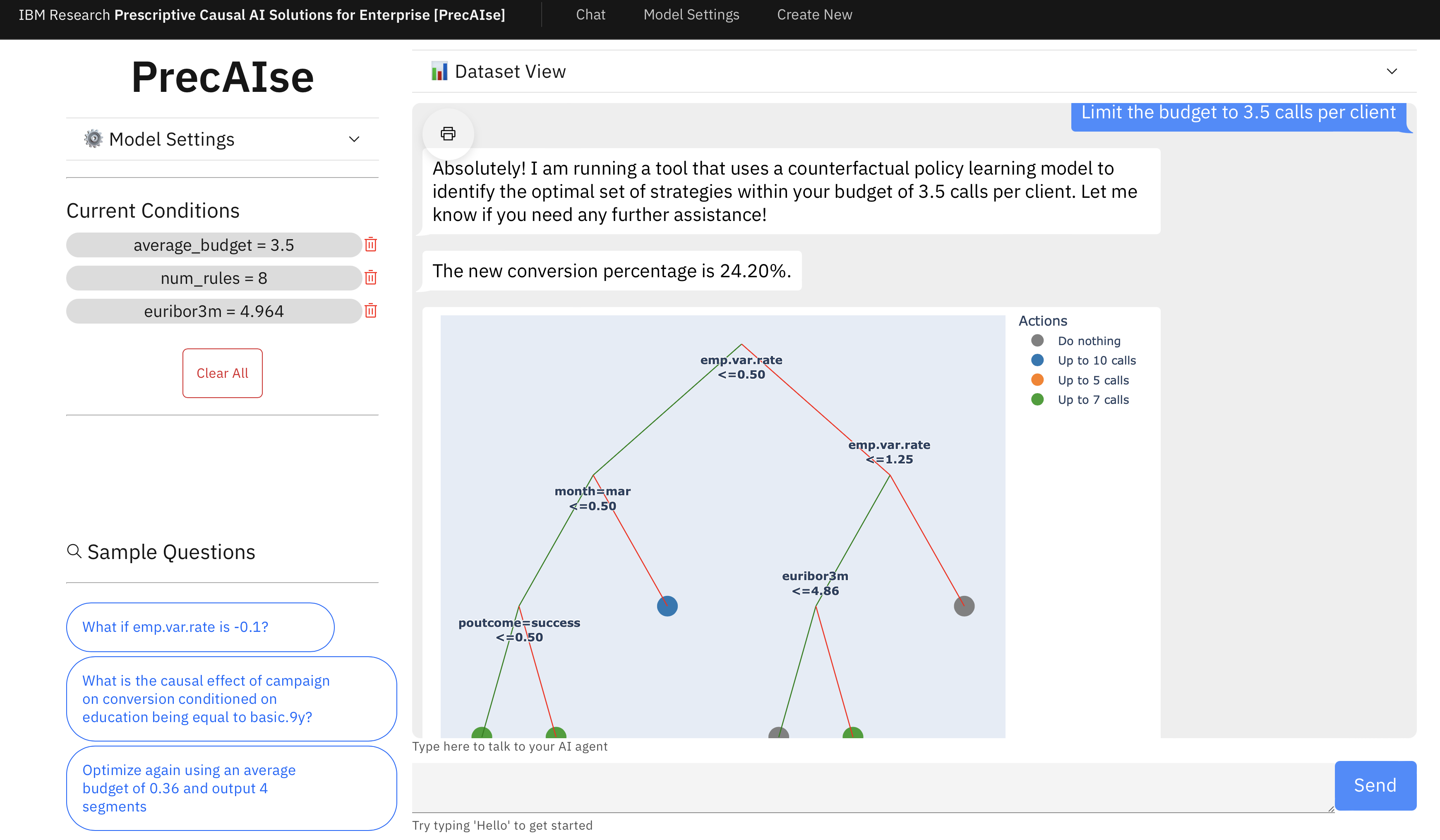}
    \caption{Optimization tool: prescriptive tree}
    \label{fig:opt1}
\end{figure}

After providing the average budget of $3.5$ calls per customer, the agent is able to run the \textit{run\_optimize} tool and produce a prescriptive tree that allows the telemarketer to decide what action to take when certain feature conditions are met, as shown in Figure \ref{fig:opt1}. The leaf nodes of this tree represent the action for the telemarketer to make. This tool also returns an updated strategy plot, mapping out how often each action should be taken on average in order to achieve the improved conversion rate of $24.20\%$.

\begin{figure}[H]
    \centering
    \includegraphics[width=0.8\linewidth]{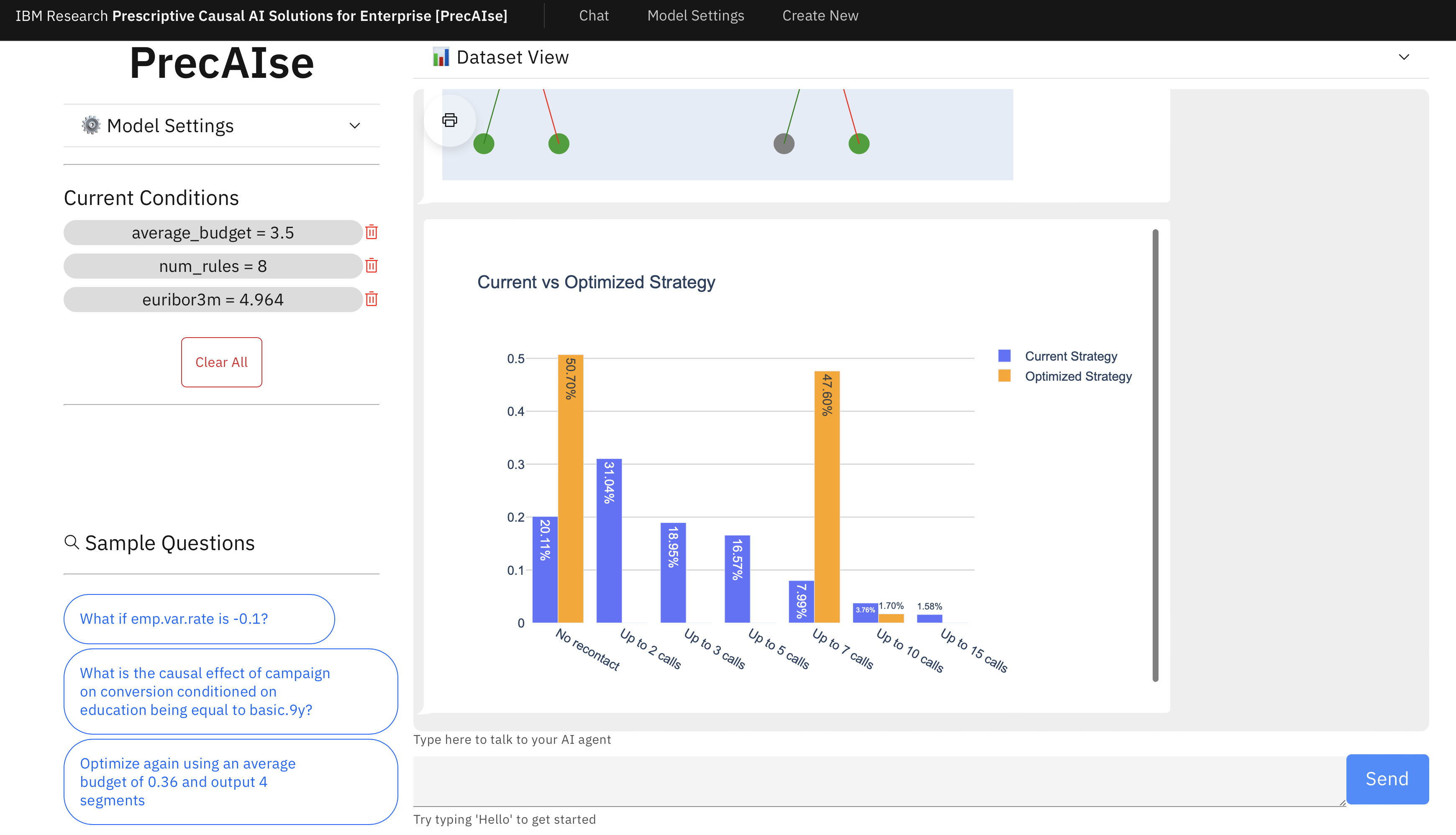}
    \caption{Optimization tool: new strategy}
    \label{fig:opt2}
\end{figure}






\bibliographystyle{plainnat}
\bibliography{precaise_main.bib}

\end{document}